\documentclass{article}

\usepackage[accepted]{icml2026}

\makeatletter
\def\addcontentsline#1#2#3{%
  \addtocontents{#1}{\protect\contentsline{#2}{#3}{\thepage}{}}}
\makeatother

\usepackage{scott}
\definecolor{lightpurple}{RGB}{168, 141, 201}

\theoremstyle{plain}

\theoremstyle{definition}

\theoremstyle{remark}

\icmltitlerunning{The Surprising Difficulty of Search in Model-Based Reinforcement Learning}

\begin{document}

\twocolumn[
  \icmltitle{The Surprising Difficulty of Search in Model-Based Reinforcement Learning}

  \icmlsetsymbol{equal}{*}

  \begin{icmlauthorlist}
    \icmlauthor{Wei-Di Chang}{equal,meta,mcgill}
    \icmlauthor{Mikael Henaff}{meta}
    \icmlauthor{Brandon Amos}{meta}
    \icmlauthor{Gregory Dudek}{mcgill}
    \icmlauthor{Scott Fujimoto}{equal,meta}
  \end{icmlauthorlist}

  \icmlaffiliation{meta}{Meta FAIR}
  \icmlaffiliation{mcgill}{McGill University, Montr\'eal, Canada}

  \icmlcorrespondingauthor{Wei-Di Chang}{wei-di.chang@mail.mcgill.ca}
  \icmlcorrespondingauthor{Scott Fujimoto}{sfujimoto@meta.com}

  \icmlkeywords{Model-Based Reinforcement Learning, World Models, Search, MPC, Reinforcement Learning, Machine Learning, ICML}

  \vskip 0.3in
]

\printAffiliationsAndNotice{\icmlEqualContribution} %

\begin{abstract}
This paper investigates search in model-based reinforcement learning~(RL). Conventional wisdom holds that long-term predictions and compounding errors are the primary obstacles for model-based RL. We challenge this view, showing that search is not a drop-in replacement for a learned policy. Surprisingly, we find that search can harm performance even when the model is highly accurate. Instead, we show that mitigating overestimation bias matters more than improving model or value function accuracy. Building on this insight, we identify that taking the minimum over an ensemble of value functions effectively addresses this bias and enables effective search, achieving state-of-the-art performance across multiple popular benchmark domains. Code can be found at \url{https://github.com/facebookresearch/MRSQ}. 
\end{abstract}

\section{Introduction}

Model-based reinforcement learning (MBRL) has long been regarded as a promising paradigm for sample-efficient decision-making in complex environments. By learning an explicit dynamics model, agents can simulate future trajectories, plan ahead, and make informed decisions without requiring millions of real-world interactions~\citep{sutton1991dyna}. 

Despite its conceptual appeal, MBRL has often struggled to meaningfully outperform its model-free counterparts \citep{van2019use, schwarzer2023bigger, fujimoto2025towards}. A common explanation for this shortfall is model accuracy and compounding errors~\citep{talvitie2014model, venkatraman2015improving, asadi2018lipschitz, lambert2022investigating}. When a model is used to simulate trajectories over long horizons, small prediction errors accumulate, leading to increasingly unreliable state estimates. This phenomenon has motivated a substantial body of research focused on improving model accuracy~\citep{oh2015action, nagabandi2018neural, hafner2019learning}, uncertainty-aware models~\citep{deisenroth2011pilco, chua2018deep, janner2019trust, henaff2019model}, and long-horizon prediction~\citep{oh2017value, silver2017predictron, lambert2021learning, janner2022planning, ma2024transformer, farebrother2025temporal}. The implicit assumption underlying these efforts is that better models will directly translate to better planning and, consequently, better performance.

{\parfillskip=0pt
In this work, we challenge this assumption by advancing a core hypothesis: \textit{model accuracy alone does not determine the effectiveness of search.} We explore this hypothesis through the lens of MR.Q~\citep{fujimoto2025towards}, a state-of-the-art algorithm that learns a dynamics model purely for representation learning, without incorporating search.
\par}

We begin in \Cref{sec:longhorizon} by establishing a counterintuitive result: search can fail even with a \textit{perfect} dynamics model and value function. Beyond a critical planning horizon, the search space expands dramatically, rendering exhaustive search computationally infeasible and high-value trajectories vanishingly rare under sampling-based approaches. Critically, this limitation is intrinsic to the search process; improving model quality cannot resolve it.

We next ask whether these theoretical limitations appear in practice. Our empirical analysis in \Cref{sec:acc_and_perf} corroborates this theoretical limitation. Despite MR.Q's architectural similarity to TD-MPC2~\citep{hansen2024td}, a leading search-based MBRL method, we find that naively incorporating search \textit{harms} MR.Q's performance. This occurs even though MR.Q's learned dynamics are comparably accurate to those of TD-MPC2, which benefits substantially from search. These results confirm that model accuracy alone does not determine whether search will be effective, but they also raise a natural question: if not model accuracy, what \textit{does} determine the success of search?

We argue that the primary obstacle lies in the interaction between search and value learning. Because exhaustive search is infeasible, practitioners rely on sampling-based methods guided by a learned value function. Due to the computational cost of search, this value function is typically trained using rollouts from a learned, non-search policy network~\citep{hansen2024td}. In \Cref{sec:overestimation}, we show that this creates a mismatch between data collected with search and the distribution assumed during training, leading to overestimation bias that degrades both value function quality and overall performance~\citep{fujimoto2019off}. 

Rather than trying to build better models, we directly address the overestimation bias introduced by search by using the minimum over an ensemble of value functions to produce pessimistic estimates for out-of-distribution actions. We instantiate this approach in a new algorithm: Model-based Representations for Search and Q-learning (MRS.Q). We evaluate MRS.Q on over 50 tasks spanning multiple benchmark domains using a single fixed set of hyperparameters. Our results demonstrate consistent improvements over both state-of-the-art model-free and model-based methods, validating that addressing overestimation, rather than solely improving model quality, is key to unlocking the benefits of search.

Our findings show the bottleneck is not just the model, but how we use it. Progress lies not only in building better dynamics models, but also in understanding and managing the fundamental tensions between value learning and search.

\section{Related Work}

{\parfillskip=0pt
\textbf{MBRL.} MBRL approaches can be broadly categorized by how they utilize their model. The simplest approach trains a model-free RL algorithm on synthetic samples generated by the model~\citep{atkeson1997comparison, abbeel2006using, ha2018world, janner2019trust}. A related strategy integrates the learned model into value learning by generating multi-step returns~\citep{oh2017value, feinberg2018model, buckman2018sample, hafner2019learning, hafner2023mastering, amos2021model}. The PILCO family of methods~\citep{deisenroth2011pilco, gal2016improving, higuera2018synthesizing} takes a different approach, using gradients from the model to directly improve the policy.
\par}

{\parfillskip=0pt
\textbf{Search in MBRL.} A more sophisticated approach uses the world model for search. One common strategy, rooted in optimal control, is model predictive control (MPC), which repeatedly searches through the model to identify the best action at each time step~\citep{draeger1995model, levine2013guided, mordatch2014combining, watter2015embed, finn2016deep, gu2016continuous, ebert2017self, banijamali2018robust, nagabandi2018neural, chua2018deep, lowrey2019plan, henaff2019model}. For discrete action spaces, Monte Carlo tree search offers a related alternative~\citep{schrittwieser2020mastering, schrittwieser2021online, ye2021mastering, wang2024efficientzero}. The current state of the art for MPC-based methods is TD-MPC2~\citep{hansen2022temporal, hansen2024td}, which uses short-horizon search with a value function. 
\par}

\textbf{Challenges in MBRL.} MBRL faces several fundamental challenges. Compounding error in auto-regression predictions limits the ability to plan over long horizons~\citep{talvitie2014model, venkatraman2015improving, asadi2018lipschitz, lambert2022investigating}. The objective mismatch problem~\citep{lambert2020objective}: models are trained to optimize immediate predictions but are ultimately evaluated by policy performance. \citet{palenicek2023diminishing} showed that oracle models offer diminishing performance gains. MR.Q~\citep{fujimoto2025towards} demonstrates that models can improve representation learning without search, raising the question of when and how search adds value. Our empirical observations reinforce these concerns, suggesting that how the model is used matters more than its accuracy. Recent extensions of TD-MPC2 have also highlighted a performance gap between the learned policy network and MPC, applying regularization to align the policy with MPC-selected actions~\citep{lin2025td, wang2025bootstrapped, zhan2025bootstrap}. \citet{lin2025td} demonstrate that MPC causes the value function to overestimate the performance of its policy network. Our experiments reveal that this failure mode extends beyond TD-MPC2, and that the value function can also overestimate the online, MPC-based policy.

\textbf{Overestimation bias.} Overestimation bias is a well-known problem in RL, stemming from the maximization of an approximate value function~\citep{thrun1993bias, hasselt2010double, DoubleDQN, fujimoto2018addressing}. In offline RL, overestimation bias is caused by errors induced by the distribution shift between the offline dataset and the learned policy~\citep{fujimoto2019off}. A similar dynamic arises when search selects actions outside the distribution seen during value function training~\citep{lin2025td, wang2025bootstrapped, zhan2025bootstrap}. 
A common strategy for addressing overestimation bias is through the minimum of value functions~\citep{fujimoto2018addressing, lan2020maxmin, an2021uncertainty}. In our work, we show that this technique is particularly effective for mitigating the overestimation induced by search.

\section{Background} \label{sec:background}

\textbf{Reinforcement learning (RL).} We follow the standard formulation~\citep{suttonbarto}, where an agent interacts with an environment modeled as a Markov Decision Process (MDP) defined by the tuple $(S, A, p, R, \gamma)$, with state space $S$, action space $A$, dynamics function $p$, reward function $R$, and discount factor $\gamma$. The objective is to learn a policy $\pi$ mapping state $s \in S$ to action $a \in A$, that maximizes the expected discounted return $\mathbb{E}_\pi\left[\sum_{t=0}^{\infty} \gamma^t r_t \right]$ of rewards $r_t \sim R(s_t, a_t)$. The value function~$Q^\pi(s,a) = \mathbb{E}_\pi\left[\sum_{t=0}^{\infty} \gamma^t r_t \right | s_0=s, a_0=a]$ measures the expected return when starting in state~$s$ and taking action~$a$. 

{\parfillskip=0pt
Model-based reinforcement learning (MBRL) approaches learn an explicit dynamics model of the environment that predicts the next state~$s_{t+1}$ given the current state~$s_t$ and action~$a_t$. In this paper, we focus on methods that use these learned models to plan via search. Search refers to the process of evaluating candidate action sequences through model rollouts to select actions. Model predictive control (MPC) is an online search framework that optimizes actions over a finite horizon using model rollouts, executing only the first action before re-planning at each time step. \par}

\textbf{TD-MPC2.} TD-MPC2~\citep{hansen2024td} is a MBRL algorithm that learns a latent world model that is used for MPC. The state embedding $\mathbf{z}_s$ is learned together with the dynamics function, reward model, and value function by co-optimizing loss functions for each:
\begin{equation} \label{eqn:TDMPC2_update}
\begin{aligned}
    \Loss ( \mathbf{z}_{s}, &F, R, Q) = \Loss_\text{Dynamics}( F(\mathbf{z}_s, a) - \mathbf{z}_{s'} ) \\
    +~&\Loss_\text{Reward}( R(\mathbf{z}_s, a) - r ) \\
    +~&\Loss_\text{Value} ( Q(\mathbf{z}_s,a) - ( r + \y Q(\mathbf{z}_{s'}, \pi(\mathbf{z}_{s'})) ) ).
\end{aligned}  
\end{equation} 
During MPC, trajectories $\tau = (\tilde{\mathbf{z}}_0, a_0, \tilde{\mathbf{z}}_1, a_1, ..., \tilde{\mathbf{z}}_N)$ are generated by recursively rolling out the model from an embedding $\mathbf{z}_{s_0}$ of an initial observed state $s_0$:  
\begin{equation}
    \tilde{\mathbf{z}}_t = F(\tilde{\mathbf{z}}_{t-1}, a_{t-1}) \quad\text{ where }\quad \tilde{\mathbf{z}}_0 = \mathbf{z}_{s_0}.
\end{equation}
The value $V$ of a trajectory~$\tau$ is computed by summing a final value $Q$ with predicted rewards along the model-predicted embeddings over a finite horizon $N$:  
\begin{equation}
    V(\tau) = \sum_{t=0}^{N-1} \y^t R(\tilde{\mathbf{z}}_t, a_t) + \y^N Q(\tilde{\mathbf{z}}_{N}, \pi(\tilde{\mathbf{z}}_{N})). \label{eqn:tdmpc2_eval_traj}
\end{equation}
{\parfillskip=0pt
TD-MPC2 uses a learned policy network~$\pi$, trained to maximize value, in its value updates, but selects actions during environment interaction using Model Predictive Path Integral (MPPI) control~\citep{williams2015model}. At each time step, candidate action trajectories are initially sampled from a mixture of random and policy-derived actions, then evaluated following \Cref{eqn:tdmpc2_eval_traj}. The highest-valued trajectories  inform the next round of candidate sampling, and this process repeats. The final trajectory is selected through value-weighted sampling, and its first action is executed. \par}

\textbf{MR.Q.} MR.Q~\citep{fujimoto2025towards} is a model-free RL method that uses a model-based objective to learn a state-action embedding~$\mathbf{z}_{sa}$ which has an approximately linear relationship to the value function~$Q^\pi(s,a)$. Unlike most model-based methods, which use search or model rollouts, MR.Q's objective is solely used to learn useful representations. More formally, MR.Q's learning objective co-optimizes the embedding~$\mathbf{z}_{sa}$ with weights capturing the reward and dynamics of the environment:
\begin{equation}
\begin{aligned}
    \Loss ( \mathbf{z}_{s}, W_p, W_r) =~&\Loss_\text{Dynamics}( \mathbf{z}_{sa}^\top W_p - \mathbf{z}_{s'}) \\
    +~&\Loss_\text{Reward}( \mathbf{z}_{sa}^\top W_r - r ). 
\end{aligned}  
\end{equation} 
The value function is trained independently with a standard Bellman update~\citep{DQN, DDPG}, taking the minimum between two value functions in the target~\citep{fujimoto2018addressing}: 
\begin{equation} \label{eqn:MRQ_value}
    \Loss_\text{Value} \lp r + \y \min (Q_1, Q_2)(\mathbf{z}_{s'a'}) - Q(\mathbf{z}_{sa}) \rp, 
\end{equation}
where the action $a'$ is sampled from the policy~ $\pi(\mathbf{z}_{s'})$. The state embedding $\mathbf{z}_s$ is learned upstream, end-to-end with $\mathbf{z}_{sa}$, i.e., $g(\mathbf{z}_s, a) = \mathbf{z}_{sa}$. Since $\mathbf{z}_{sa}$ is a non-linear function of $\mathbf{z}_s$ and $a$, the resulting dynamics and reward function are functionally similar to TD-MPC2's. Crucially, although MR.Q leverages a model-based objective, it is strictly used to learn the embeddings rather than forward dynamics predictions or search.

\textbf{Adding and removing MPC.} The architectural similarity between TD-MPC2 and MR.Q allows us to isolate the effect of search by adding MPC to MR.Q and removing it from TD-MPC2. For clarity, let $\pi_\text{MPC}$ denote the implicit policy induced by MPC, which selects actions via MPPI using the learned model and value function (\Cref{eqn:tdmpc2_eval_traj}). Methods with MPC select actions according to $\pi_\text{MPC}$, while methods without MPC select actions directly from the learned policy~$\pi$. 

\textbf{Distribution shift.} Distribution shift arises in value estimation when the behavior policy used for data collection differs from the target policy $\pi$ used in the value update (\Cref{eqn:TDMPC2_update,eqn:MRQ_value}). For example, if MPC selects actions during data collection but the value function is trained with actions from the learned policy~$\pi$, this mismatch may cause the value function to be queried on actions it was not trained on, ultimately leading to unreliable value estimates~\citep{fujimoto2019off}.

\section{The Challenge of Search}

In this section, we investigate the fundamental challenges limiting the effectiveness of search in MBRL. Conventional knowledge suggests that model accuracy is the key predictor of MBRL performance. We challenge this assumption, demonstrating that it does not strictly hold. While model accuracy remains important, our results suggest that there are other underappreciated factors at play.

We present three key findings:
\begin{enumerate}[topsep=-1pt, itemsep=1pt, leftmargin=10pt]
    \item \textbf{Naive search can fail with a perfect model.} We show theoretically that even with perfect model and value functions, uniform random search can fail with high probability over longer horizons due to the exponential growth of the search space~(\Cref{sec:longhorizon}).

    \item {\parfillskip=0pt \textbf{Model accuracy alone does not determine whether search improves performance.} Contrary to expectations, adding MPC to MR.Q~\citep{fujimoto2025towards} degrades its performance, despite MR.Q having a more accurate model than TD-MPC2~\citep{hansen2024td}, a method where MPC proves beneficial~(\Cref{sec:acc_and_perf}). \par}

    \item \textbf{Search induces overestimation bias.} We show that simply replacing the learned policy with MPC when collecting data introduces overestimation bias that degrades value function quality~(\Cref{sec:overestimation}).
\end{enumerate}
Together, these findings motivate the need for methods that explicitly address these challenges rather than focusing solely on improving model accuracy. 

\subsection{Search with a perfect model} \label{sec:longhorizon}

\begin{figure}
\centering
\small
\begin{tikzpicture}[
indata/.style={draw,circle, minimum size=30pt, very thick},
outdata/.style={draw=none,circle, minimum size=30pt, very thick}
]
\node[indata] (s0) at (0,0) {$s_0$};
\node[indata] (s1) at (2,0) {$s_1$};
\node[indata] (s2) at (4,0) {$s_2$};
\node[outdata] (s3) at (6,0) {\hspace{-16pt} \dots}; 
\node[indata, fill=gray!10] (term) at (2,-2) {$s_\text{absorb}$};

\node[outdata] (spacing) at (-1,0) {}; 

\draw[-{Latex[length=5pt]}, thick] (s0) -- node[above]{$a_0$} (s1);
\draw[-{Latex[length=5pt]}, thick] (s1) -- node[above]{$a_0$} (s2);
\draw[-{Latex[length=5pt]}, thick] (s2) -- node[above]{$a_0$} (s3);

\draw[-{Latex[length=5pt]}, thick] (s0) -- node[left]{$\{a_i\}$~~} (term);
\draw[-{Latex[length=5pt]}, thick] (s1) -- node[left]{$\{a_i\}$} (term);
\draw[-{Latex[length=5pt]}, thick] (s2) -- node[right]{~~$\{a_i\}$} (term);

\draw[-{Latex[length=5pt]}, thick] (term.150) arc (50:260+50:10pt) node[midway,left]{$a_0, \{a_i\}$};

\end{tikzpicture}
\caption{\textbf{Absorbing N-chain.} The absorbing N-chain is a simple environment with $N+1$ states and $A$ actions. The action $a_0$ moves the agent forward and any other action $\{a_i\}$ sends the agent to a self-looping absorbing state $s_\text{absorb}$. The reward is $0$ everywhere, other than the final state~$s_{N-1}$ of the N-chain.
} \label{fig:nchain}
\vspace{-4pt}
\end{figure}
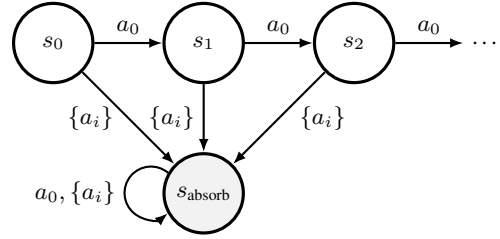

We begin by examining the theoretical limits of search under ideal conditions. Perhaps surprisingly, search can fail with high probability even with perfect dynamics and a perfect value function. To illustrate this, consider the N-chain MDP defined in \Cref{fig:nchain}. In this environment, a single action~$a_0$ advances the agent along the chain, while all other actions lead to an absorbing state. The reward is zero everywhere except at the final state. We define search failure as the inability to discover any trajectory with non-zero value, and analyze a uniform random search that samples action sequences, simulates trajectories via the model, and selects the trajectory with the highest value.

In this N-chain environment, regular states have zero reward but non-zero value, since the final state remains accessible. In contrast, the absorbing state has both zero reward and zero value. Consequently, if the agent selects any action other than~$a_0$, the trajectory will end in the absorbing state and will necessarily have zero value. Therefore, even with access to the ground-truth value function, an agent can only select action~$a_0$ at each time step to discover a trajectory with non-zero value. 

Under random search, the probability of finding such a trajectory is $1 - (1 - \frac{1}{A^n})^m$, where $A$ is the number of actions, $n$ is the search horizon, and $m$ is the number of sampled trajectories. Intuitively, only 1 of $A^n$ possible action sequences consists entirely of $a_0$, so the probability of randomly sampling a successful trajectory decreases exponentially with horizon length.

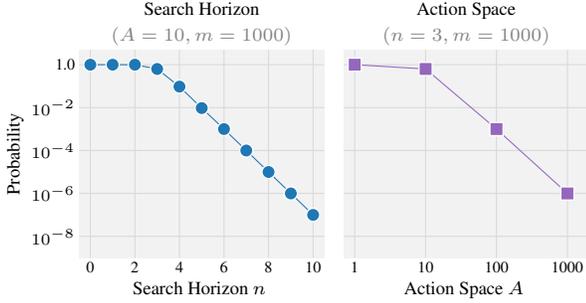
\begin{figure}[t]
    \begin{tikzpicture}[trim axis right]
        \begin{axis}[
            height=0.1615\textwidth,
            width=0.19\textwidth,
            title={\shortstack{Search Horizon\vphantom{p}\\\textcolor{gray}{$(A=10, m=1000)$}}},
            ylabel={Probability},
            xlabel={Horizon $n$ \vphantom{$(A$p}},
            ymode=log,
            ytick={1e-8, 1e-6, 1e-4, 1e-2, 1},
            yticklabels={$10^{-8}$, $10^{-6}$, $10^{-4}$, $10^{-2}$, 1.0},
            xtick={0, 2, 4, 6, 8, 10},
            xticklabels={0, 2, 4, 6, 8, 10},
            xmin=-0.5,
            xmax=10.5,
            ymin=1e-9,
            ymax=5,
            axis background/.style={fill=gray!10},
            grid=major,
            major grid style={black!15},
            tick style={draw=none},
        ]
        \addplot[sb_blue, mark=*, mark size=2.5pt, mark options={draw=white, line width=0.5pt}] coordinates {(0,1) (1,1) (2,0.999957) (3,0.632305)(4,0.0951671) (5, 0.0095022) (6, 0.000999501) (7, 0.000099995) (8, 9.99995e-6) (9, 9.999995e-7) (10, 9.9999995e-8)};
        \end{axis}
    \end{tikzpicture}
    \begin{tikzpicture}[trim axis right]
        \begin{axis}[
            height=0.1615\textwidth,
            width=0.19\textwidth,
            title={\shortstack{Action Space\\\textcolor{gray}{$(n=3, m=1000)$}}},
            xlabel={Number of actions $A$ \vphantom{$(A$p}},
            ymode=log,
            xmode=log,
            ytick={1e-8, 1e-6, 1e-4, 1e-2, 1},
            yticklabels={},
            xtick={1, 10, 100, 1000},
            xticklabels={1, 10, 100, 1000},
            xmin=0.7,
            xmax=2000,
            ymin=1e-9,
            ymax=5,
            axis background/.style={fill=gray!10},
            grid=major,
            major grid style={black!15},
            tick style={draw=none},
        ]
        \addplot[sb_purple, mark=square*, mark size=2.5pt, mark options={draw=white, line width=0.5pt}] coordinates {(1,1) (10, 0.632305) (100, 0.000999500666125591) (1000, 9.999995005e-7)};
        \end{axis}
    \end{tikzpicture}

    \captionof{figure}{\textbf{Probability of finding a non-zero value trajectory.} We analyze how the probability of discovering a non-zero value trajectory varies with the search horizon length~(left) and the action space size~(right) in the absorbing N-chain~(\Cref{fig:nchain}) according to $1 - (1 - \frac{1}{A^n})^m$. This demonstrates that the probability decays rapidly as either parameter increases.
    } \label{fig:prob}
\end{figure}

In \Cref{fig:prob} we show that this probability degrades rapidly as the search horizon increases. For example, with $A = 10$ actions and $m = 1000$ trajectories, a search horizon of $n=3$ gives a probability of approximately $0.63$ of discovering a non-zero valued trajectory. However, extending the search horizon to $n=10$ causes the probability to fall to $10^{-7}$.

The failure in this example could be avoided by acting according to the value function rather than relying on search. More broadly, this illustrates that depending solely on imagined trajectories may be insufficient when the search space is large, particularly in domains with continuous states and actions. While modern search-based methods such as TD-MPC2~\citep{hansen2024td} use policy-guided sampling to narrow the search space, they still face combinatorial challenges as the horizon increases, particularly when the policy is imperfect or the reward landscape requires precise action sequences. This helps explain why MBRL methods often employ short search horizons (three time steps), both to constrain the search space and to reduce compounding model error.

\subsection{Does model accuracy determine performance?} \label{sec:acc_and_perf}

Having established that search can fail even with perfect models in theoretical settings, we now examine the role of model accuracy in practical settings with learned components. We compare two recent state-of-the-art methods, MR.Q~\citep{fujimoto2025towards} and TD-MPC2~\citep{hansen2024td}, both with and without MPC. We focus on these methods because the similarities in architecture and learning objectives allow us to isolate the effect of search from confounding factors.

In \Cref{table:acc_perf} we evaluate each method's learned dynamics model accuracy and task performance using the following metrics:

\begin{itemize}[topsep=-1pt, itemsep=1pt,leftmargin=10pt]
\item 
\textbf{Dynamics error.} The mean-squared error between predicted and actual next-state embeddings, weighted by $\y^t$ over the three-step search horizon. To ensure comparable error magnitudes, we modify MR.Q's state encoder by replacing its final activation function with Simplicial Embeddings (SEM) \citep{lavoie2023simplicial}, matching TD-MPC2's architecture and ensuring that differences in embedding scale do not inflate one method's error.

\item 
\textbf{Unroll error.} The absolute error between the value prediction used in MPC and the ground-truth trajectory value. We compute the value prediction by unrolling the dynamics over the search horizon, and summing the discounted predicted rewards with the value estimate of the final predicted state~(\Cref{eqn:tdmpc2_eval_traj}). The ground-truth value is computed similarly, unrolling trajectories with the simulator and summing the true discounted reward over the full episode length.  

\item 
\textbf{Performance $\Delta$.} The performance difference with and without MPC (i.e., MR.Q with MPC vs. MR.Q, and TD-MPC2 vs. TD-MPC2 without MPC). Positive values indicate that MPC improved performance.
\end{itemize}

\begin{table}[t]
\footnotesize
\centering
\setlength{\tabcolsep}{4pt}
\caption{\textbf{Comparing model accuracy and performance.} Dynamics error and unroll error do not correlate with MPC performance. For dynamics error and unroll error, we highlight which method performs \hlfancy{sb_blue!10}{better} or \hlfancy{sb_purple!30}{worse}. Performance $\Delta$ indicates whether MPC \hlfancy{sb_green!20}{improved} or \hlfancy{sb_red!30}{harmed} each method's performance.} \label{table:acc_perf}
\vspace{-3pt}
\begin{tabular}{l@{\hspace{3pt}}lrrr}
\toprule
& Environment & Dyn.\@ Error & Unroll Error & Perf. $\Delta$ \\
\midrule
\multirow{17}{*}{\rotatebox{90}{MR.Q}} 
& acrobot-swingup & \cellcolor{sb_blue!10} 3.62e-07 & \cellcolor{sb_blue!10} 2.98 & \cellcolor{sb_red!30} -107.41 \\
& cheetah-run & \cellcolor{sb_blue!10} 3.16e-05 & \cellcolor{sb_blue!10} 2.26 & \cellcolor{sb_red!30} -173.53 \\
& dog-stand & \cellcolor{sb_purple!30} 8.59e-05 & \cellcolor{sb_purple!30} 15.53 & \cellcolor{sb_red!30} -238.35 \\
& dog-walk & \cellcolor{sb_blue!10} 7.36e-05 & \cellcolor{sb_purple!30} 11.13 & \cellcolor{sb_red!30} -260.71 \\
& dog-trot & \cellcolor{sb_blue!10} 2.56e-05 & \cellcolor{sb_blue!10} 3.32 & \cellcolor{sb_red!30} -207.10 \\
& dog-run & \cellcolor{sb_blue!10} 3.51e-05 & \cellcolor{sb_blue!10} 1.14 & \cellcolor{sb_red!30} -208.12 \\
& hopper-stand & \cellcolor{sb_blue!10} 2.48e-05 & \cellcolor{sb_blue!10} 3.09 & \cellcolor{sb_red!30} -27.87 \\
& hopper-hop & \cellcolor{sb_blue!10} 3.77e-05 & \cellcolor{sb_purple!30} 14.11 & \cellcolor{sb_red!30} -26.83 \\
& humanoid-stand & \cellcolor{sb_purple!30} 5.30e-05 & \cellcolor{sb_blue!10} 3.98 & \cellcolor{sb_red!30} -756.61 \\
& humanoid-walk & \cellcolor{sb_purple!30} 1.22e-04 & \cellcolor{sb_blue!10} 14.93 & \cellcolor{sb_red!30} -291.00 \\
& humanoid-run & \cellcolor{sb_blue!10} 4.18e-05 & \cellcolor{sb_blue!10} 1.87 & \cellcolor{sb_red!30} -95.71 \\
& walker-run & \cellcolor{sb_blue!10} 3.04e-05 & \cellcolor{sb_purple!30} 12.51 & \cellcolor{sb_red!30} -263.36 \\
& \textcolor{gray}{(Gym)} Ant & \cellcolor{sb_blue!10} 4.78e-05 & \cellcolor{sb_blue!10} 136.90 & \cellcolor{sb_green!20} 166.23 \\
& \textcolor{gray}{(Gym)} HalfCheetah & \cellcolor{sb_blue!10} 5.39e-05 & \cellcolor{sb_purple!30} 207.99 & \cellcolor{sb_red!30} -8395.10 \\
& \textcolor{gray}{(Gym)} Hopper & \cellcolor{sb_purple!30} 3.63e-05 & \cellcolor{sb_blue!10} 12.81 & \cellcolor{sb_red!30} -1990.26 \\
& \textcolor{gray}{(Gym)} Humanoid & \cellcolor{sb_blue!10} 5.23e-05 & \cellcolor{sb_purple!30} 809.28 & \cellcolor{sb_red!30} -5692.61 \\
& \textcolor{gray}{(Gym)} Walker2d & \cellcolor{sb_blue!10} 2.47e-05 & \cellcolor{sb_blue!10} 36.60 & \cellcolor{sb_red!30} -5869.88 \\
\midrule
\multirow{17}{*}{\rotatebox{90}{TD-MPC2}} 
& acrobot-swingup & \cellcolor{sb_purple!30} 1.39e-06 & \cellcolor{sb_purple!30} 7.15 & \cellcolor{sb_green!20} 121.01 \\
& cheetah-run & \cellcolor{sb_purple!30} 4.85e-05 & \cellcolor{sb_purple!30} 3.37 & \cellcolor{sb_green!20} 40.04 \\
& dog-stand & \cellcolor{sb_blue!10} 4.36e-05 & \cellcolor{sb_blue!10} 3.68 & \cellcolor{sb_red!30} -348.53 \\
& dog-walk & \cellcolor{sb_purple!30} 1.00e-04 & \cellcolor{sb_blue!10} 2.67 & \cellcolor{sb_green!20} 2.71 \\
& dog-trot & \cellcolor{sb_purple!30} 8.21e-05 & \cellcolor{sb_purple!30} 69.30 & \cellcolor{sb_green!20} 80.41 \\
& dog-run & \cellcolor{sb_purple!30} 1.25e-04 & \cellcolor{sb_purple!30} 5.33 & \cellcolor{sb_green!20} 23.25 \\
& hopper-stand & \cellcolor{sb_purple!30} 2.12e-04 & \cellcolor{sb_purple!30} 33.89 & \cellcolor{sb_green!20} 236.21 \\
& hopper-hop & \cellcolor{sb_purple!30} 1.61e-04 & \cellcolor{sb_blue!10} 6.72 & \cellcolor{sb_green!20} 202.32 \\
& humanoid-stand & \cellcolor{sb_blue!10} 3.37e-05 & \cellcolor{sb_purple!30} 41.22 & \cellcolor{sb_green!20} 234.44 \\
& humanoid-walk & \cellcolor{sb_blue!10} 8.83e-05 & \cellcolor{sb_purple!30} 40.62 & \cellcolor{sb_green!20} 273.34 \\
& humanoid-run & \cellcolor{sb_purple!30} 9.33e-05 & \cellcolor{sb_purple!30} 9.77 & \cellcolor{sb_green!20} 66.86 \\
& walker-run & \cellcolor{sb_purple!30} 4.42e-05 & \cellcolor{sb_blue!10} 12.21 & \cellcolor{sb_green!20} 157.90 \\
& \textcolor{gray}{(Gym)} Ant & \cellcolor{sb_purple!30} 1.15e-04 & \cellcolor{sb_purple!30} 538.92 & \cellcolor{sb_red!30} -1900.55 \\
& \textcolor{gray}{(Gym)} HalfCheetah & \cellcolor{sb_purple!30} 1.59e-04 & \cellcolor{sb_blue!10} 59.74 & \cellcolor{sb_green!20} 6501.50 \\
& \textcolor{gray}{(Gym)} Hopper & \cellcolor{sb_blue!10} 1.46e-05 & \cellcolor{sb_purple!30} 211.65 & \cellcolor{sb_red!30} -783.52 \\
& \textcolor{gray}{(Gym)} Humanoid & \cellcolor{sb_purple!30} 2.44e-04 & \cellcolor{sb_blue!10} 304.54 & \cellcolor{sb_red!30} -351.25 \\
& \textcolor{gray}{(Gym)} Walker2d & \cellcolor{sb_purple!30} 3.32e-05 & \cellcolor{sb_purple!30} 43.56 & \cellcolor{sb_red!30} -1963.40 \\
\bottomrule
\end{tabular}
\end{table}

Dynamics and unroll errors directly measure model quality. Unroll error, in particular, measures the quality of the signal that MPC uses to select actions (\Cref{eqn:tdmpc2_eval_traj}), as it reflects both dynamics and value prediction accuracy. Performance~$\Delta$ shows the change in performance using the \textit{identical} MPC procedure with each algorithm.

Despite comparable model accuracy between methods, using MPC gives divergent outcomes, improving TD-MPC2 while degrading MR.Q. \Cref{table:acc_perf} quantifies this surprising asymmetry. Although both methods exhibit similar dynamics and value prediction errors, TD-MPC2's performance improves substantially with MPC across nearly all DM Control Suite (DMC) tasks~\citep{tassa2018deepmind}, while MR.Q's consistently decreases. Given their architectural and loss function similarities, we would expect comparable error rates to translate into comparable search benefits. 

The substantial gap in MPC performance confirms that model accuracy alone does not determine whether search will help, suggesting other factors are at play. 

\input{figures/value_error-v6}

\subsection{Why does search harm MR.Q?} \label{sec:overestimation}

Given that model accuracy does not fully explain these results, we now investigate what other factors are responsible. Following \citet{lin2025td, wang2025bootstrapped, zhan2025bootstrap}, we hypothesize that distribution shift and its downstream effects on value estimation are the key bottleneck to search performance.

Consider the standard value function update used by both TD-MPC2 and MR.Q, based on a value target that evaluates a learned policy network~$\pi$:
\begin{equation} \label{eqn:td_update}
Q(s,a) \approx r + \y Q(s', \pi(s')).
\end{equation}
This value update evaluates the learned policy $\pi$ rather than the action originally taken in the environment. When MPC is used to collect data, there is a mismatch in which the value function is trained on actions from $\pi_\text{MPC}$ but queried on actions sampled from $\pi$. Because $\pi \neq \pi_\text{MPC}$, these queries are out-of-distribution, which prior work has shown leads to approximation errors and overestimated values \citep{fujimoto2019off}.

One might resolve this by using search to compute value targets directly, but doing so is prohibitively expensive. Every sample in a mini-batch requires a target, multiplying the number of costly search calls by the batch size and update-to-data ratio (e.g., $256\times$ with default hyperparameters).

To test this hypothesis, we measure the accuracy of value functions learned by MR.Q, MR.Q with MPC, TD-MPC2, and our proposed approach, MRS.Q (introduced in the following section) in \Cref{fig:value_error}. Value error is computed by taking the difference between the learned value estimate and the true discounted return of the behavior policy: the learned policy network for MR.Q, and the MPC policy~$\pi_\text{MPC}$ for the others. Unlike prior work~\citep{lin2025td} which examined overestimation bias of the learned policy~$\pi$, our experiments focus on the behavior policy (largely the MPC policy~$\pi_\text{MPC}$).

\Cref{fig:value_error} shows that simply changing the behavior policy from the learned policy (MR.Q) to MPC (MR.Q+MPC) introduces significant overestimation bias. Moreover, \Cref{fig:value_error} reveals a clear relationship between value error and the performance change $\Delta$ from using MPC (\Cref{table:acc_perf}):
\begin{itemize}[topsep=-1pt, itemsep=1pt,leftmargin=10pt]
    \item \textbf{MR.Q+MPC:} Switching from the learned policy to MPC introduces overestimation in nearly every environment. MPC improves performance only in Ant, where overestimation is the lowest.
    \item \textbf{TD-MPC2:} Compared to MRQ+MPC, overestimation is considerably lower overall. However, it remains high in environments where TD-MPC2's learned policy outperforms vanilla TD-MPC2 (dog-stand, Ant, Hopper, Humanoid, Walker2d).
\end{itemize}
These findings suggest that (1) MPC introduces significant overestimation bias, and (2) although the design choices in TD-MPC2 help mitigate this bias, they are insufficient as a standalone solution.

\section{Effective Search in Model-based RL} \label{sec:MRSQ}

Having identified overestimation bias as a key challenge limiting search effectiveness, we now present our approach. Our algorithm, \textbf{M}odel-based \textbf{R}epresentations for \textbf{S}earch and \textbf{Q}-learning (MRS.Q), builds on MR.Q with targeted modifications that enable effective use of search.

MRS.Q introduces three key modifications to MR.Q: (1) search-based action selection with a short horizon, (2) aggressive minimization over an ensemble of value functions, and (3) minor hyperparameter changes. %

\textbf{Action selection.} Following TD-MPC2, MRS.Q selects actions using Model Predictive Path Integral (MPPI) control~\citep{williams2015model} when acting in the environment. As such, we also use the same MPPI hyperparameters as TD-MPC2. Critically, this means that we adopt the same short MPC horizon of $3$ steps, using the value function to estimate returns beyond this window. As discussed in \Cref{sec:longhorizon}, naive search can fail with high probability when planning over long horizons due to the expanded search space. A short horizon keeps the search space tractable and limits compounding error in the model rollout.

\begin{figure}[t]
\centering
\begin{minipage}{0.02\textwidth}
\rotatebox{90}{\parbox{150pt}{\centering \scriptsize Squared difference in selected action}}%
\end{minipage}\begin{minipage}{0.5\textwidth}
\raggedright
\hspace{-6pt}
\begin{tikzpicture}[trim axis right]
    \begin{axis}[
        height=0.084\textwidth,
        width=0.42\textwidth,
        title={\vphantom{p}acrobot-swingup},
        ylabel={},
        xlabel={},
        xtick={-10, 90, 190, 290, 390, 490},
        xticklabels={},
        ytick={0.0, 2.2016689607040054},
        yticklabels={0.0, 2.2},
    ]
    \addplot graphics [
    ymin=-0.05, ymax=2.311752408739206,
    xmin=-15.0, xmax=495.0,
    ]{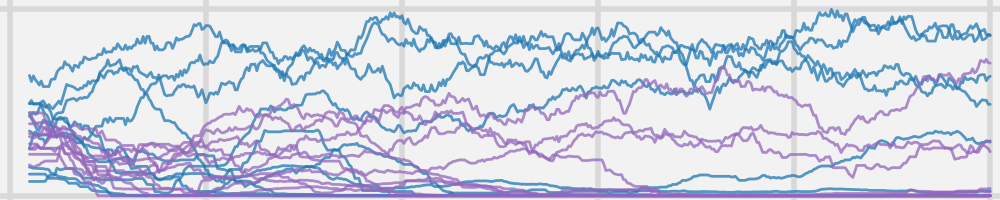};
    \end{axis}
\end{tikzpicture}
\hspace{-6pt}
\begin{tikzpicture}[trim axis right]
    \begin{axis}[
        height=0.084\textwidth,
        width=0.42\textwidth,
        title={\vphantom{p}cheetah-run},
        ylabel={},
        xlabel={},
        xtick={-10, 90, 190, 290, 390, 490},
        xticklabels={},
        ytick={0.0, 0.46960455551743513},
        yticklabels={0.0, 0.5},
    ]
    \addplot graphics [
    ymin=-0.05, ymax=0.4930847832933069,
    xmin=-15.0, xmax=495.0,
    ]{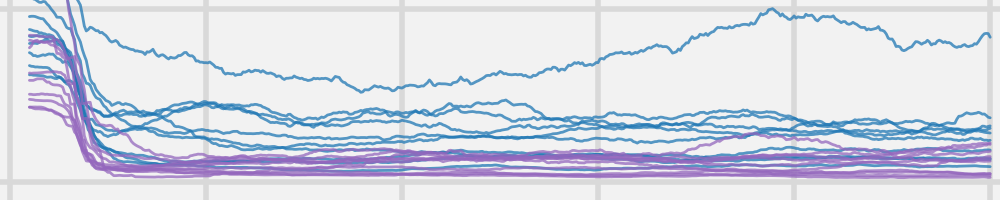};
    \end{axis}
\end{tikzpicture}

\vspace{-6pt}
\hspace{-6pt}
\begin{tikzpicture}[trim axis right]
    \begin{axis}[
        height=0.084\textwidth,
        width=0.42\textwidth,
        title={\vphantom{p}dog-run},
        ylabel={},
        xlabel={},
        xtick={-10, 90, 190, 290, 390, 490},
        xticklabels={},
        ytick={0.0, 0.3745756033062934},
        yticklabels={0.0, 0.4},
    ]
    \addplot graphics [
    ymin=-0.05, ymax=0.3933043834716081,
    xmin=-15.0, xmax=495.0,
    ]{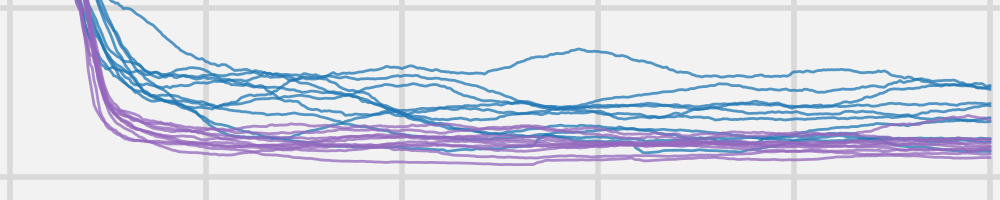};
    \end{axis}
\end{tikzpicture}
\hspace{-6pt}
\begin{tikzpicture}[trim axis right]
    \begin{axis}[
        height=0.084\textwidth,
        width=0.42\textwidth,
        title={\vphantom{p}dog-stand},
        ylabel={},
        xlabel={},
        xtick={-10, 90, 190, 290, 390, 490},
        xticklabels={},
        ytick={0.0, 0.3556630152463913},
        yticklabels={0.0, 0.4},
    ]
    \addplot graphics [
    ymin=-0.05, ymax=0.3734461660087109,
    xmin=-15.0, xmax=495.0,
    ]{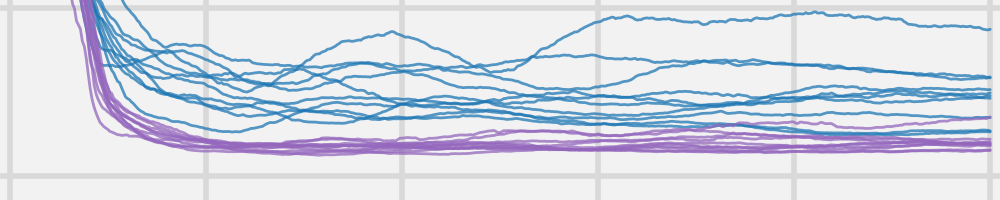};
    \end{axis}
\end{tikzpicture}

\vspace{-6pt}
\hspace{-6pt}
\begin{tikzpicture}[trim axis right]
    \begin{axis}[
        height=0.084\textwidth,
        width=0.42\textwidth,
        title={\vphantom{p}dog-trot},
        ylabel={},
        xlabel={},
        xtick={-10, 90, 190, 290, 390, 490},
        xticklabels={},
        ytick={0.0, 0.30992041319608626},
        yticklabels={0.0, 0.3},
    ]
    \addplot graphics [
    ymin=-0.05, ymax=0.3254164338558906,
    xmin=-15.0, xmax=495.0,
    ]{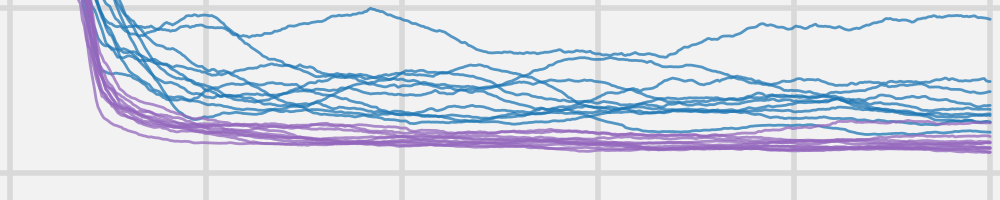};
    \end{axis}
\end{tikzpicture}
\hspace{-6pt}
\begin{tikzpicture}[trim axis right]
    \begin{axis}[
        height=0.084\textwidth,
        width=0.42\textwidth,
        title={\vphantom{p}dog-walk},
        ylabel={},
        xlabel={},
        xtick={-10, 90, 190, 290, 390, 490},
        xticklabels={},
        ytick={0.0, 0.4560314583778379},
        yticklabels={0.0, 0.5},
    ]
    \addplot graphics [
    ymin=-0.05, ymax=0.47883303129672977,
    xmin=-15.0, xmax=495.0,
    ]{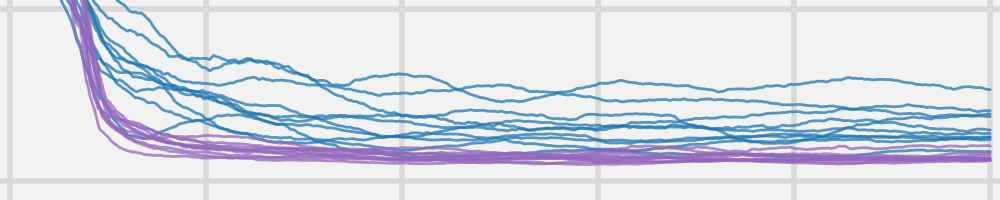};
    \end{axis}
\end{tikzpicture}

\vspace{-6pt}
\hspace{-6pt}
\begin{tikzpicture}[trim axis right]
    \begin{axis}[
        height=0.084\textwidth,
        width=0.42\textwidth,
        title={\vphantom{p}hopper-hop},
        ylabel={},
        xlabel={},
        xtick={-10, 90, 190, 290, 390, 490},
        xticklabels={},
        ytick={0.0, 0.8726341897249229},
        yticklabels={0.0, 0.9},
    ]
    \addplot graphics [
    ymin=-0.05, ymax=0.9162658992111691,
    xmin=-15.0, xmax=495.0,
    ]{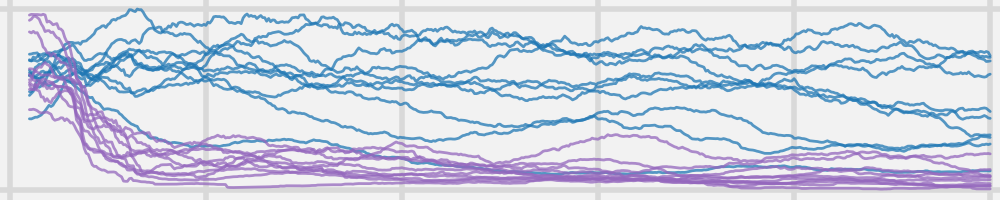};
    \end{axis}
\end{tikzpicture}
\hspace{-6pt}
\begin{tikzpicture}[trim axis right]
    \begin{axis}[
        height=0.084\textwidth,
        width=0.42\textwidth,
        title={\vphantom{p}hopper-stand},
        ylabel={},
        xlabel={},
        xtick={-10, 90, 190, 290, 390, 490},
        xticklabels={},
        ytick={0.0, 0.8825626486539839},
        yticklabels={0.0, 0.9},
    ]
    \addplot graphics [
    ymin=-0.05, ymax=0.9266907810866831,
    xmin=-15.0, xmax=495.0,
    ]{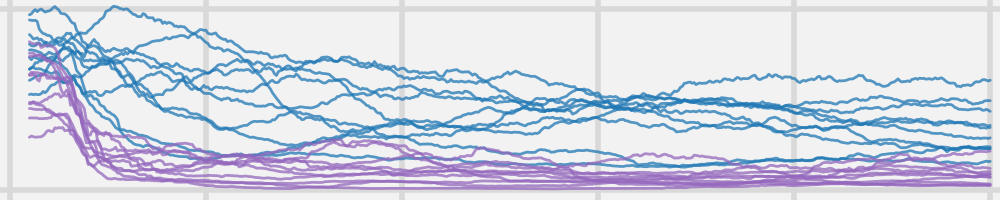};
    \end{axis}
\end{tikzpicture}

\vspace{-6pt}
\hspace{-6pt}
\begin{tikzpicture}[trim axis right]
    \begin{axis}[
        height=0.084\textwidth,
        width=0.42\textwidth,
        title={\vphantom{p}humanoid-run},
        ylabel={},
        xlabel={},
        xtick={-10, 90, 190, 290, 390, 490},
        xticklabels={},
        ytick={0.0, 0.7984412020444861},
        yticklabels={0.0, 0.8},
    ]
    \addplot graphics [
    ymin=-0.05, ymax=0.8383632621467104,
    xmin=-15.0, xmax=495.0,
    ]{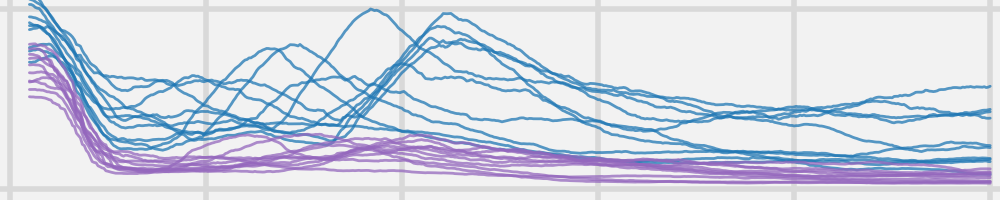};
    \end{axis}
\end{tikzpicture}
\hspace{-6pt}
\begin{tikzpicture}[trim axis right]
    \begin{axis}[
        height=0.084\textwidth,
        width=0.42\textwidth,
        title={\vphantom{p}humanoid-stand},
        ylabel={},
        xlabel={},
        xtick={-10, 90, 190, 290, 390, 490},
        xticklabels={},
        ytick={0.0, 0.7320483714342118},
        yticklabels={0.0, 0.7},
    ]
    \addplot graphics [
    ymin=-0.05, ymax=0.7686507900059224,
    xmin=-15.0, xmax=495.0,
    ]{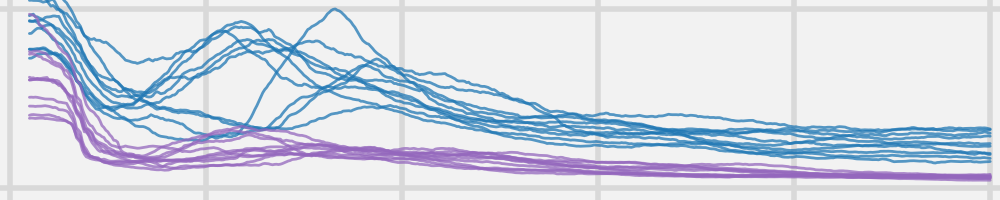};
    \end{axis}
\end{tikzpicture}

\vspace{-6pt}
\hspace{-6pt}
\begin{tikzpicture}[trim axis right]
    \begin{axis}[
        height=0.084\textwidth,
        width=0.42\textwidth,
        title={\vphantom{p}humanoid-walk},
        ylabel={},
        xlabel={Time steps (1M)},
        xtick={-10, 90, 190, 290, 390, 490},
        xticklabels={0.0, 0.2, 0.4, 0.6, 0.8, 1.0},
        ytick={0.0, 0.8028322851657868},
        yticklabels={0.0, 0.8},
    ]
    \addplot graphics [
    ymin=-0.05, ymax=0.8429738994240761,
    xmin=-15.0, xmax=495.0,
    ]{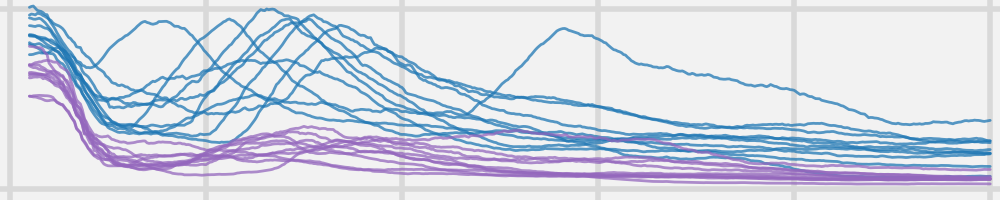};
    \end{axis}
\end{tikzpicture}
\hspace{-6pt}
\begin{tikzpicture}[trim axis right]
    \begin{axis}[
        height=0.084\textwidth,
        width=0.42\textwidth,
        title={\vphantom{p}walker-run},
        ylabel={},
        xlabel={Time steps (1M)},
        xtick={-10, 90, 190, 290, 390, 490},
        xticklabels={0.0, 0.2, 0.4, 0.6, 0.8, 1.0},
        ytick={0.0, 0.569753158390522},
        yticklabels={0.0, 0.6},
    ]
    \addplot graphics [
    ymin=-0.05, ymax=0.5982408163100481,
    xmin=-15.0, xmax=495.0,
    ]{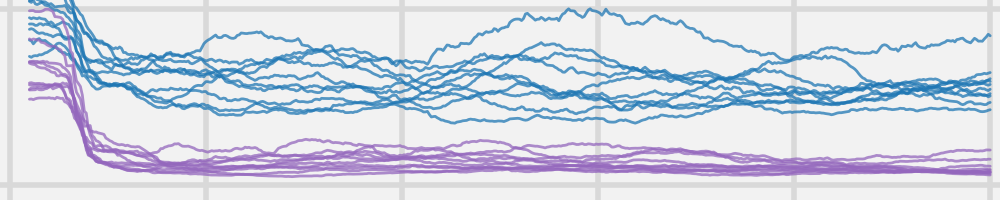};
    \end{axis}
\end{tikzpicture}
\end{minipage}

\centering
\fcolorbox{gray!10}{gray!10}{
\small
\cnotdashed{sb_blue} MPC \quad
\cnotdashed{sb_purple} Policy network
}

\captionof{figure}{\textbf{Change in selected action.} We measure how actions selected by MPC and the policy network change over the course of learning. Specifically, we compute the mean-squared difference between selected actions at intervals of 1000 time steps, with all actions scaled to $[-1,1]$. Across a wide range of environments, actions selected by MPC change significantly more than those selected by the policy network.}
\label{fig:policy_change}
\end{figure}

\textbf{Minimum over value functions.} Recent work addressing distribution shift in MBRL~\citep{lin2025td, wang2025bootstrapped, zhan2025bootstrap} has focused on constraining the policy to match actions selected by search. However, this approach is problematic: search-selected actions shift rapidly during training as the model evolves. 

In \Cref{fig:policy_change}, we compare the rate of change of actions selected by search versus the learned policy network over the course of learning. Throughout training, search-selected actions exhibit substantially more variation than policy network actions, suggesting that constraining the policy to match such an unstable target may destabilize learning or introduce noise into training. 

Rather than constraining or modifying the learned policy, we address distribution shift by directly reducing overestimation in the value function. We achieve this through aggressive minimization across an ensemble of 10 value functions. While taking the minimum over a pair of value functions has become a standard technique~\citep{fujimoto2018addressing}, this approach naturally extends to larger ensembles:
\begin{equation} \label{eqn:min}
Q(s,a) \approx r + \y \min_{i \in \{1,2,..,10\}} Q_i(s', \pi(s')).
\end{equation}
Crucially, we apply this minimization wherever the value function is evaluated with the policy: not only when computing value targets (as above), but also when computing final values during MPC trajectory evaluation (\Cref{eqn:tdmpc2_eval_traj}). Applying the minimum during trajectory evaluation prevents the search procedure from favoring potentially inflated value estimates over model-predicted rewards, while ensuring consistency in how the value function is used throughout the algorithm.

TD-MPC2 also employs multiple value functions ($N=5$), however, rather than using the full ensemble, it randomly samples a pair, taking the minimum for value updates~\citep{chen2020randomized} and the mean during MPC. As shown in \Cref{fig:value_error}, using the minimum across the full ensemble significantly reduces overestimation compared to MR.Q ($N=2$) naively combined with MPC. We further defend this choice in the ablation studies in \Cref{sec:ablations}. 

\begin{figure*}[t]
    \centering    
    \begin{tikzpicture}[trim axis right]
        \begin{axis}[
            height=0.1615\textwidth,
            width=0.19\textwidth,
            title={\vphantom{p}Gym},
            ylabel={TD3-Normalized},
            xlabel={Timesteps (1M)},
            xtick={0.0, 40.0, 80.0, 120.0, 160.0, 200.0},
            xticklabels={0.0, 0.2, 0.4, 0.6, 0.8, 1.0},
            ytick={0.0, 0.4, 0.8, 1.2, 1.6},
            yticklabels={0.0, 0.4, 0.8, 1.2, 1.6},
        ]
        \addplot graphics [
        ymin=-0.08247154577500702, ymax=1.6482933749126056,
        xmin=-10.0, xmax=210.0,
        ]{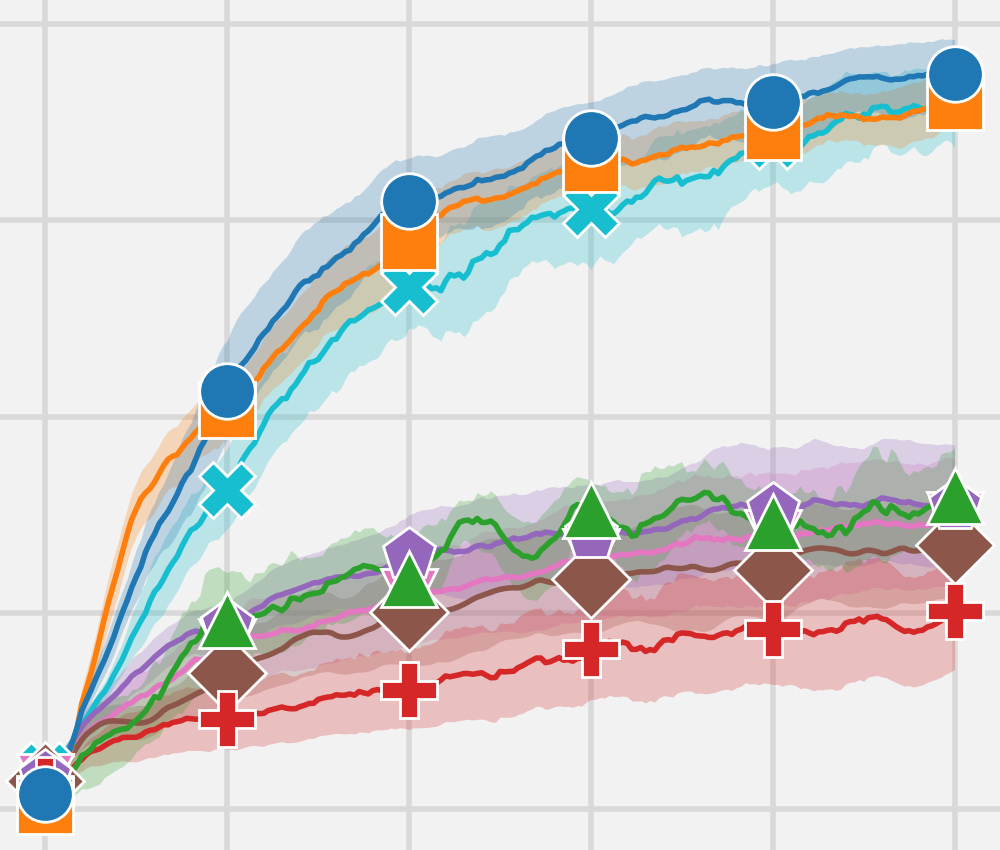};
        \end{axis}
    \end{tikzpicture}
    \begin{tikzpicture}[trim axis right]
        \begin{axis}[
            height=0.1615\textwidth,
            width=0.19\textwidth,
            title={DMC\vphantom{p}},
            ylabel={Total Reward (1k)},
            xlabel={Time steps (1M)},
            xtick={0.0, 20.0, 40.0, 60.0, 80.0, 100.0, 120.0},
            xticklabels={0.0, 0.2, 0.4, 0.6, 0.8, 1.0},
            ytick={0.0, 2.0, 4.0, 6.0, 8.0, 10.0},
            yticklabels={0.0, 0.2, 0.4, 0.6, 0.8, 1.0},
        ]
        \addplot graphics [
        ymin=-0.5, ymax=9.50,
        xmin=-5.0, xmax=105.0,
        ]{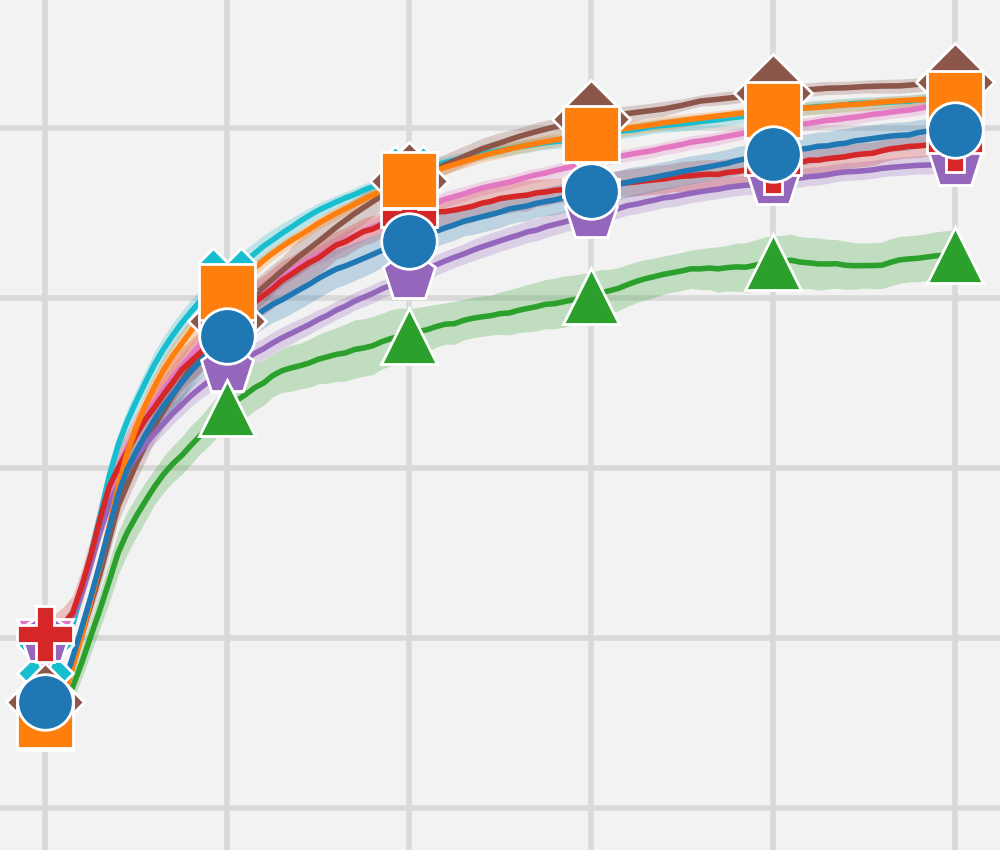};
        \end{axis}
    \end{tikzpicture}
    \begin{tikzpicture}[trim axis right]
        \begin{axis}[
            height=0.1615\textwidth,
            width=0.19\textwidth,
            title={HumanoidBench - No Hand\vphantom{p}},
            xlabel={Time steps (1M)},
            xtick={0.0, 40.0, 80.0, 120.0, 160.0, 200.0},
            xticklabels={0.0, 0.2, 0.4, 0.6, 0.8, 1.0},
            ytick={0.0, 0.1, 0.2, 0.3, 0.4, 0.5, 0.6},
            yticklabels={0.0, 0.1, 0.2, 0.3, 0.4, 0.5, 0.6, 0.7},
        ]
        \addplot graphics [
        ymin=-0.05, ymax=0.65,
        xmin=-10.0, xmax=210.0,
        ]{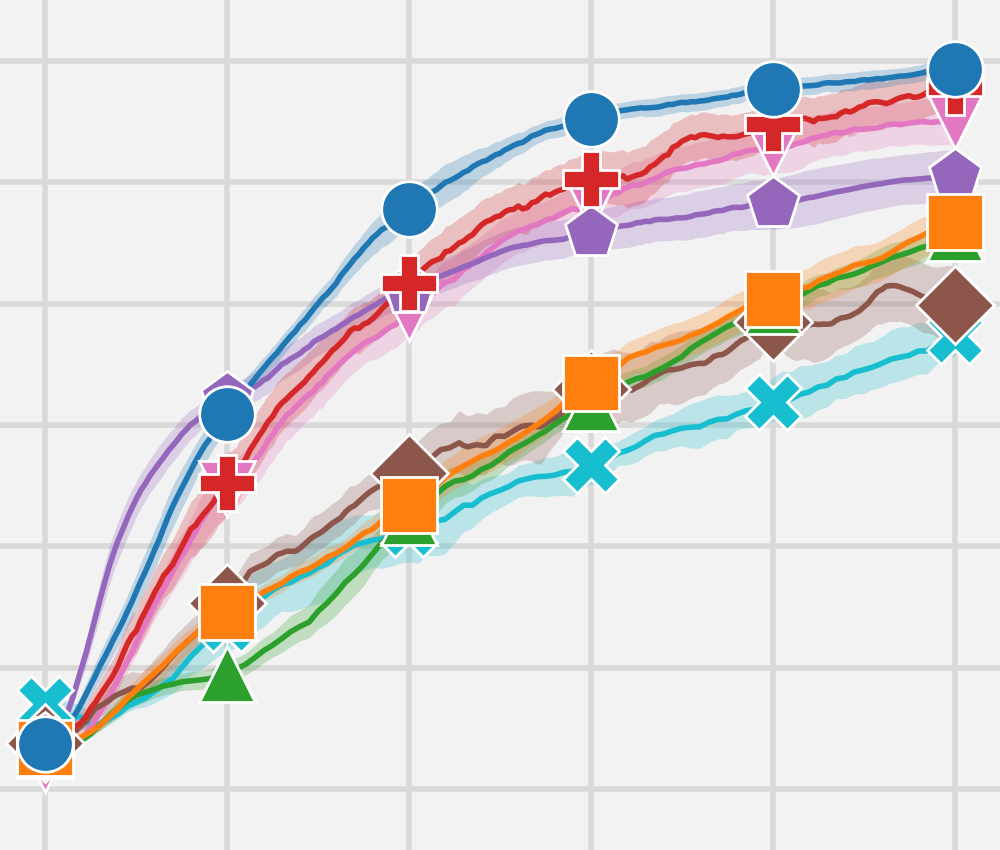};
        \end{axis}
    \end{tikzpicture}
    \begin{tikzpicture}[trim axis right]
        \begin{axis}[
            height=0.1615\textwidth,
            width=0.19\textwidth,
            title={HumanoidBench - Hand\vphantom{p}},
            xlabel={Time steps (1M)},
            xtick={0.0, 40.0, 80.0, 120.0, 160.0, 200.0},
            xticklabels={0.0, 0.2, 0.4, 0.6, 0.8, 1.0},
            ytick={0.0, 0.1, 0.2, 0.3, 0.4, 0.5, 0.6},
            yticklabels={0.0, 0.1, 0.2, 0.3, 0.4, 0.5, 0.6, 0.7},
        ]
        \addplot graphics [
        ymin=-0.05, ymax=0.65,
        xmin=-10.0, xmax=210.0,
        ]{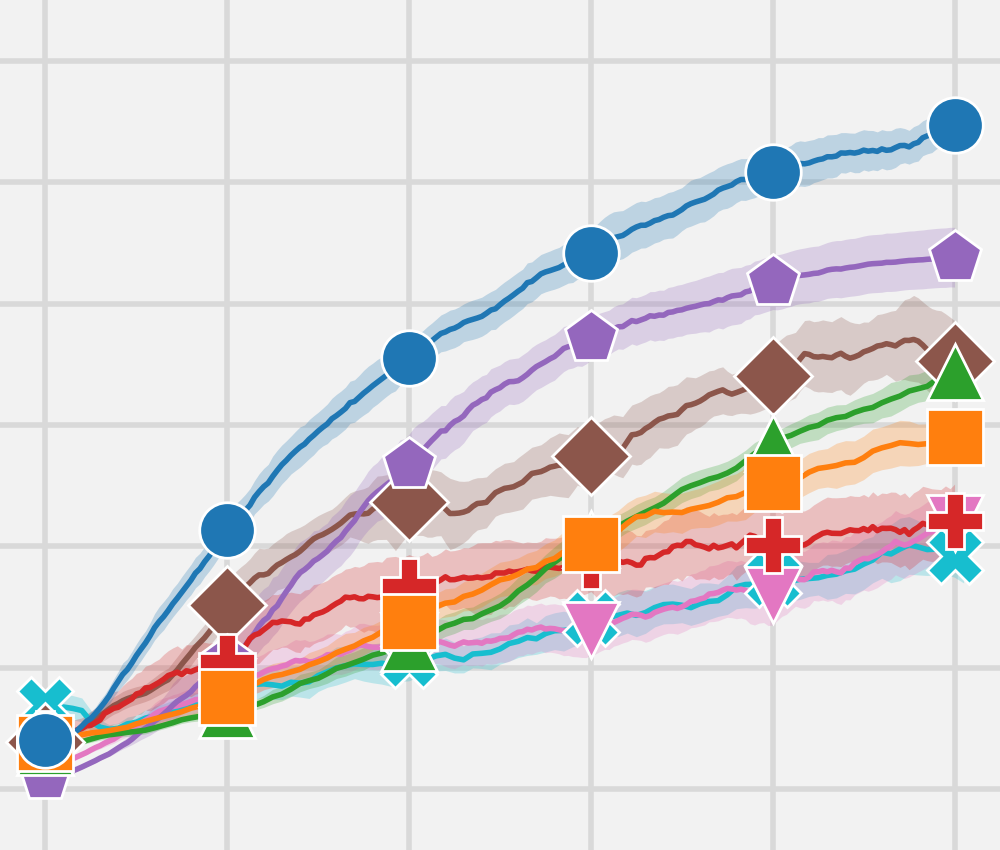};
        \end{axis}
    \end{tikzpicture}

    \fcolorbox{gray!10}{gray!10}{
    \small
    \colorcircle{white}{sb_blue} MRS.Q (Ours) \quad
    \colorsquare{white}{sb_orange} MR.Q \quad
    \colortriangle{white}{sb_green} MR.Q + MPC \quad
    \colorplus{white}{sb_red} TD-MPC2 \quad
    \colorpoly{white}{sb_purple} TD-M(PC)$^2$ \quad
    \colordiamond{white}{sb_brown}  BMPC \quad
    \colorinvertedtriangle{white}{sb_pink} BOOM \quad
    \colorx{white}{sb_cyan} SimbaV2 
    }
    
    \captionof{figure}{\textbf{Aggregate learning curves.} Our approach MRS.Q offers significant performance benefits over state-of-the-art model-based RL methods. The shaded area captures a 95\% stratified bootstrap confidence interval across 10 seeds. Gym scores are normalized using TD3 performance values~\citep{fujimoto2018addressing} (see \Cref{appendix:envs} for details).} \label{fig:agg_curves}
\end{figure*}

\begin{table*}[ht]

    \caption{\textbf{Aggregate Results.} Average final performance across each benchmark at 1M time steps. \textcolor{gray}{[Bracketed values]} indicate 95\% stratified bootstrap confidence intervals over 10 seeds. Gym scores are normalized to TD3 performance; other domains are scaled by dividing by 1k. The \hlfancy{sb_blue!25}{highest score} per benchmark is highlighted.}
        \label{table:results}
        \centering
        \footnotesize
        \setlength{\tabcolsep}{0pt}
        \begin{tabular}{@{\hspace{6pt}}l@{\hspace{12pt}} c@{\hspace{12pt}} r@{}l@{\hspace{14pt}} r@{}l@{\hspace{14pt}} r@{}l@{\hspace{14pt}} r@{}l@{\hspace{6pt}}}
        \toprule
        Algorithm & MPC &\multicolumn{2}{c}{Gym}  & \multicolumn{2}{c}{DMC} & \multicolumn{2}{c}{HB (No Hand)} & \multicolumn{2}{c}{HB (Hand)} \\
        \midrule
        MR.Q & \textcolor{sb_red}{$\times$} 
        & 1.46 & ~\textcolor{gray}{[1.41, 1.52]} 
        & 0.84 & ~\textcolor{gray}{[0.83, 0.84]}
        & 0.48 & ~\textcolor{gray}{[0.46, 0.49]}
        & 0.31 & ~\textcolor{gray}{[0.29, 0.32]}
        \\
        MR.Q + MPC & \textcolor{sb_green}{\checkmark}
        & 0.67 & ~\textcolor{gray}{[0.55, 0.88]}
        & 0.65 & ~\textcolor{gray}{[0.63, 0.68]}
        & 0.46 & ~\textcolor{gray}{[0.45, 0.48]}
        & 0.38 & ~\textcolor{gray}{[0.37, 0.39]}
        \\ 
        TD-MPC2 & \textcolor{sb_green}{\checkmark}
        & 0.41 & ~\textcolor{gray}{[0.27, 0.57]}
        & 0.78 & ~\textcolor{gray}{[0.77, 0.80]}
        & 0.58 & ~\textcolor{gray}{[0.56, 0.60]}
        & 0.22 & ~\textcolor{gray}{[0.19, 0.25]}
        \\ 
        TD-M(PC)$^2$ & \textcolor{sb_green}{\checkmark}
        & 0.62 & ~\textcolor{gray}{[0.50, 0.74]}
        & 0.76 & ~\textcolor{gray}{[0.75, 0.77]}
        & 0.51 & ~\textcolor{gray}{[0.48, 0.53]}
        & 0.44 & ~\textcolor{gray}{[0.41, 0.46]}
        \\ 
        BMPC & \textcolor{sb_green}{\checkmark}
        & 0.54 & ~\textcolor{gray}{[0.44, 0.63]} 
        & \cellcolor{sb_blue!25} 0.86 & \cellcolor{sb_blue!25}~\textcolor{gray}{[0.86, 0.87]}
        & 0.40 & ~\textcolor{gray}{[0.37, 0.43]}
        & 0.38 & ~\textcolor{gray}{[0.35, 0.40]}
        \\
        BOOM & \textcolor{sb_green}{\checkmark}
        & 0.61 & ~\textcolor{gray}{[0.47, 0.74]}
        & 0.83 & ~\textcolor{gray}{[0.83, 0.84]}
        & 0.55 & ~\textcolor{gray}{[0.53, 0.57]}
        & 0.23 & ~\textcolor{gray}{[0.20, 0.26]}
        \\ 
        SimbaV2 & \textcolor{sb_red}{$\times$} 
        & 1.44 & ~\textcolor{gray}{[1.35, 1.52]}
        & 0.84 & ~\textcolor{gray}{[0.83, 0.85]}
        & 0.38 & ~\textcolor{gray}{[0.36, 0.40]}
        & 0.18 & ~\textcolor{gray}{[0.17, 0.20]}
        \\
        \midrule
        MRS.Q & \textcolor{sb_green}{\checkmark}
        & \cellcolor{sb_blue!25} 1.54 & \cellcolor{sb_blue!25}~\textcolor{gray}{[1.46, 1.60]}
        & 0.81 & ~\textcolor{gray}{[0.79, 0.82]}
        & \cellcolor{sb_blue!25} 0.59 & \cellcolor{sb_blue!25}~\textcolor{gray}{[0.58, 0.60]} 
        & \cellcolor{sb_blue!25} 0.58 & \cellcolor{sb_blue!25}~\textcolor{gray}{[0.57, 0.58]}
        \\ 
        \bottomrule
        \end{tabular}
    \end{table*}

\textbf{Hyperparameter changes.} In addition to the ensemble, we make only three hyperparameter changes to MR.Q:
\begin{itemize}[topsep=-1pt, itemsep=1pt, leftmargin=10pt]
    \item \textbf{No exploration noise.} As shown in \Cref{fig:policy_change}, search produces greater variation in action selection than the policy network. This variation in action selection implicitly induces exploration, making additive Gaussian noise unnecessary. We therefore remove it entirely (see \Cref{sec:ablations}).
    
    \item \textbf{Simplicial embedding (SEM).} Following TD-MPC2, we apply SEM~\citep{lavoie2023simplicial, obando2025simplicial} to both the state encoder and the output of the dynamics model. SEM normalizes latent representations onto the probability simplex, which stabilizes multi-step dynamics rollouts by reducing drift in predicted states. Because SEM produces sparse representations with smaller loss magnitudes, we increase the dynamics loss weight from $1 \rightarrow 20$ to match the value used by TD-MPC2 and balance the loss terms. 
    
    \item \textbf{Increased terminal loss weight.} Since predicting when episodes terminate is essential for accurate value estimation, we increase the weight of the terminal loss function from MR.Q's default of $0.1 \rightarrow 1$. 
\end{itemize}

All other hyperparameters and design choices follow the original MR.Q defaults. We highlight these minor adjustments to show that enabling effective search in MR.Q does not require major modifications to hyperparameter settings.

\section{Results}
\label{sec:experiments}

We evaluate MRS.Q on a diverse set of continuous control tasks to demonstrate the performance benefits of MRS.Q. 

{\parfillskip=0pt
\textbf{Benchmarks.} We evaluate on over 50 tasks spanning three benchmark domains: MuJoCo~\citep{mujoco, towers2024gymnasium}, DeepMind Control Suite (DMC)~\citep{tassa2018deepmind}, and HumanoidBench~\citep{sferrazza2024humanoidbench}. For HumanoidBench, we split the results into two subsets, one which uses hand features, matching the experimental setting used by BMPC and BOOM~\citep{wang2025bootstrapped, zhan2025bootstrap}, and one which does not, matching the experimental setting used by SimbaV2~\citep{lee2025hyperspherical}. \par}

\textbf{Baselines.} We evaluate MRS.Q against several methods, including MR.Q~\citep{fujimoto2025towards} (both with and without MPC), TD-MPC2~\citep{hansen2024td}, three state-of-the-art extensions of TD-MPC2 that regularize the policy to mitigate distribution shift: TD-M(PC)$^2$~\citep{lin2025td}, BMPC~\citep{wang2025bootstrapped}, and BOOM~\citep{zhan2025bootstrap}, and the state-of-the-art model-free method SimbaV2~\citep{lee2025hyperspherical}. All methods are trained for 1M environment steps (full details in \Cref{app:baselines}). 

\textbf{Evaluation protocol.} For each task, we train agents for 1M environment steps and 10 seeds. We evaluate agents every 5k steps for 10 episodes without exploration noise. We report mean performance with 95\% confidence intervals.

\subsection{Main Results}

We evaluate aggregate performance across all benchmarks, as shown in \Cref{fig:agg_curves} and \Cref{table:results}. MRS.Q outperforms all baselines on three out of four benchmarks and remains competitive on the fourth (DMC). This demonstrates the effectiveness of our proposed modifications. While naively combining MPC with MR.Q degrades performance, our approach enables MRS.Q to fully leverage MPC.

Although TD-MPC2 was specifically designed for search and both methods use the same search procedure, MRS.Q achieves substantially stronger results, particularly on Gym and the higher-dimensional HumanoidBench tasks with hand features. This demonstrates that MRS.Q successfully unlocks MR.Q's potential for search.

Furthermore, MRS.Q surpasses the TD-MPC2 extensions (TD-M(PC)$^2$, BMPC, BOOM) that regularize the policy towards search-selected actions. This suggests that simply imitating search behavior is insufficient to address distribution shift, likely due to the rapid evolution of selected actions during search (see \Cref{fig:policy_change}).

\begin{table*}[ht]
    \caption{\textbf{Ablation Studies.} Average difference in normalized performance from varying design choices across each benchmark. \textcolor{gray}{[Bracketed values]} indicate 95\% stratified bootstrap confidence intervals over 5 seeds. %
    Negative changes are highlighted with red, according to \hlfancy{sb_red!10}{$[-0.01,-0.2)$}, \hlfancy{sb_red!25}{$[-0.2,-0.5)$}, \hlfancy{sb_red!40}{$(\leq -0.5)$}. Positive changes are similarly highlighted with green \hlfancy{sb_green!10}{$(> 0.01)$}.}
        \label{table:ablations}
        \centering
        \footnotesize
        \setlength{\tabcolsep}{0pt}
        \begin{tabular}{@{\hspace{6pt}}l@{\hspace{14pt}} c@{\hspace{14pt}} r@{}l@{\hspace{14pt}} r@{}l@{\hspace{14pt}} r@{}l@{\hspace{14pt}} r@{}l@{\hspace{6pt}}}
        \toprule
        Ablation & Algorithm & \multicolumn{2}{c}{Gym}  & \multicolumn{2}{c}{DMC} & \multicolumn{2}{c}{HB (No Hand)} & \multicolumn{2}{c}{HB (Hand)} \\
        \midrule 
        2 Value Functions & MRS.Q
        & \cellcolor{sb_red!40} -0.63 & \cellcolor{sb_red!40}~\textcolor{gray}{[-0.72, -0.53]}
        & \cellcolor{sb_red!10} -0.04 & \cellcolor{sb_red!10}~\textcolor{gray}{[-0.05, -0.03]}
        & \cellcolor{sb_red!10} -0.18 & \cellcolor{sb_red!10}~\textcolor{gray}{[-0.2, -0.16]}
        & \cellcolor{sb_red!25} -0.40 & \cellcolor{sb_red!25}~\textcolor{gray}{[-0.41, -0.39]}
        \\ 
        5 Value Functions & MRS.Q
        & \cellcolor{sb_red!25} -0.37 & \cellcolor{sb_red!25}~\textcolor{gray}{[-0.47, -0.27]} 
        & 0.00 & ~\textcolor{gray}{[-0.01, 0.01]}
        & \cellcolor{sb_red!10} -0.02 & \cellcolor{sb_red!10}~\textcolor{gray}{[-0.03, -0.01]}
        & \cellcolor{sb_red!10} -0.13 & \cellcolor{sb_red!10}~\textcolor{gray}{[-0.14, -0.12]}
        \\ 
        20 Value Functions & MRS.Q
        & \cellcolor{sb_red!10} -0.05 & \cellcolor{sb_red!10}~\textcolor{gray}{[-0.10, -0.00]}
        & \cellcolor{sb_red!10} -0.03 & \cellcolor{sb_red!10}~\textcolor{gray}{[-0.06, -0.01]} 
        & \cellcolor{sb_green!10} 0.03 & \cellcolor{sb_green!10}~\textcolor{gray}{[0.01, 0.04]}
        & \cellcolor{sb_green!10} 0.03 & \cellcolor{sb_green!10}~\textcolor{gray}{[0.02, 0.04]}
        \\ 
        \midrule
        Exploration & MRS.Q
        & \cellcolor{sb_red!10} -0.13 & \cellcolor{sb_red!10}~\textcolor{gray}{[-0.24, -0.06]}
        & \cellcolor{sb_red!10} -0.02 & \cellcolor{sb_red!10}~\textcolor{gray}{[-0.04, -0.00]}
        & \cellcolor{sb_red!10} -0.02 & \cellcolor{sb_red!10}~\textcolor{gray}{[-0.03, -0.01]}
        & \cellcolor{sb_red!10} -0.02 & \cellcolor{sb_red!10}~\textcolor{gray}{[-0.04, -0.01]}
        \\
        SEM & MRS.Q
        & \cellcolor{sb_red!25} -0.28 & \cellcolor{sb_red!25}~\textcolor{gray}{[-0.62, 0.05]} 
        & \cellcolor{sb_red!10} -0.04 & \cellcolor{sb_red!10}~\textcolor{gray}{[-0.06, -0.01]}
        & \cellcolor{sb_green!10} 0.06 & \cellcolor{sb_green!10}~\textcolor{gray}{[0.05, 0.07]}
        & \cellcolor{sb_green!10} 0.10 & \cellcolor{sb_green!10}~\textcolor{gray}{[0.10, 0.10]}
        \\ 
        Min (Ensemble) & MRS.Q
        & \cellcolor{sb_red!25} -0.49 & \cellcolor{sb_red!25}~\textcolor{gray}{[-0.55, -0.44]}
        &  0.01 & ~\textcolor{gray}{[0.00, 0.03]}
        & \cellcolor{sb_red!10} -0.05 & \cellcolor{sb_red!10}~\textcolor{gray}{[-0.07, -0.04]}
        & \cellcolor{sb_red!10} -0.19 & \cellcolor{sb_red!10}~\textcolor{gray}{[-0.20, -0.18]}
        \\ 
        Min (MPC) & MRS.Q
        & \cellcolor{sb_red!25} -0.33 & \cellcolor{sb_red!25}~\textcolor{gray}{[-0.42, -0.19]}
        & \cellcolor{sb_red!10} -0.02 & \cellcolor{sb_red!10}~\textcolor{gray}{[-0.04, -0.00]}
        & \cellcolor{sb_red!10} -0.04 & \cellcolor{sb_red!10}~\textcolor{gray}{[-0.05, -0.03]}
        & \cellcolor{sb_red!10} -0.19 & \cellcolor{sb_red!10}~\textcolor{gray}{[-0.21, -0.18]}
        \\ 
        \midrule
        Min (10) & TD-MPC2
        & \cellcolor{sb_green!10} 0.07 & \cellcolor{sb_green!10}~\textcolor{gray}{[-0.08, 0.21]}
        & \cellcolor{sb_green!10} 0.03 & \cellcolor{sb_green!10}~\textcolor{gray}{[0.02, 0.05]}
        & \cellcolor{sb_green!10} 0.02 & \cellcolor{sb_green!10}~\textcolor{gray}{[0.00, 0.03]}
        & \cellcolor{sb_green!10} 0.11 & \cellcolor{sb_green!10}~\textcolor{gray}{[0.07, 0.15]}
        \\
        \bottomrule
        \end{tabular}
    \end{table*}

\subsection{Ablation Studies}
\label{sec:ablations}

We ablate key components of MRS.Q to understand their individual contributions. \Cref{table:ablations} summarizes the results.

\textbf{Number of value functions.} We vary the number of value functions used in the ensemble. We find that performance improves as the ensemble size increases up to approximately 10 value functions, after which gains plateau. As shown in \Cref{fig:value_error}, using the default 2 value functions is insufficient to address the overestimation induced by MPC, while an ensemble of 10 provides a good trade-off between bias reduction and computational cost.

\textbf{Exploration noise.} Re-adding exploration noise with MR.Q's default of $\N(0,0.2^2)$ induces a small performance drop across the four benchmarks. As illustrated in \Cref{fig:policy_change}, using MPC already induces substantial change in selected actions, making additional exploration noise on top of MPC unnecessary and often harmful. 

\textbf{SEM.} Replacing SEM in the state encoder with MR.Q's default ELU activation function~\citep{clevert2015fast} reduces performance on Gym and DMC, confirming that stabilizing latent dynamics is important for effective multi-step planning. However, this replacement slightly improves performance on HumanoidBench, suggesting that SEM is not the primary driver of performance gains in MRS.Q. 

{\parfillskip=0pt
\textbf{Minimum over ensemble.} Instead of taking the minimum over the ensemble of value functions, we match TD-MPC2 by taking the minimum over two randomly sampled ensemble members~\citep{chen2020randomized}. We find there is a significant performance loss in Gym and HumanoidBench - Hand. This suggests that overestimation bias mitigation is particularly important in high-dimensional tasks with early termination. Conversely, we remark this technique is not necessary in the DMC benchmark, possibly due to smoother dense rewards, and the lack of early termination. \par}

{\parfillskip=0pt
\textbf{Minimum in MPC.} Our minimization strategy replaces all evaluations of the value function with a minimum over the entire ensemble. To measure the impact of minimization within MPC, we experiment with using the mean of the ensemble when evaluating trajectories during MPC, i.e., in \cref{eqn:tdmpc2_eval_traj}, as done in TD-MPC2. We find that this gives a moderate performance drop across the four benchmarks. \par}

\textbf{Minimum TD-MPC2.} To evaluate the generalizability of our approach, we apply it to TD-MPC2 by taking the minimum over an ensemble of 10 value functions. We find that this improves performance across all four benchmarks.

\section{Conclusion}

{\parfillskip=0pt
In this work, we highlight a critical bottleneck in MBRL methods: the search process itself and the distribution shift it introduces. When value functions are trained using a non-search policy, but using search-gathered data, the resulting distribution shift leads to overestimation bias that undermines the benefits search is meant to provide. Our method, MRS.Q, adds search to MR.Q~\citep{fujimoto2025towards} and addresses this overestimation bias by taking the minimum over an ensemble of value functions. This targeted intervention gives consistent improvements over MR.Q and state-of-the-art model-based and model-free RL baselines on more than 50 tasks across popular benchmark domains. \par}

{\parfillskip=0pt
More broadly, our results challenge a foundational assumption in MBRL: better models lead to better performance. Our analysis demonstrates that search can fail despite highly accurate, or even perfect, dynamics models, and that naively incorporating search into state-of-the-art algorithms can actively degrade performance. These findings show that model accuracy alone is insufficient and equal attention should be given to how models are used. By directly addressing the challenges of search, we can unlock performance gains that better models alone cannot provide. 
\par}

\textbf{Limitations.} Despite MRS.Q's strong performance, several limitations remain. First, the computational cost of maintaining 10 value functions and running MPC at each step is substantial. Although this cost is comparable to other MBRL methods such as TD-MPC2~\citep{hansen2024td}, it represents a significant increase over the search-free MR.Q (see \Cref{sec:appendix_cost}). Second, the aggressive minimization strategy provides smaller gains on tasks with dense rewards and no early termination (e.g., DMC), suggesting that the approach is most beneficial in settings where overestimation is severe. Lastly, because we adopt TD-MPC2's search method, our approach inherits its short planning horizon, and our evaluation is limited to continuous control benchmarks. Nevertheless, our work offers novel insights into MBRL and highlights potential pitfalls of incorporating search into existing algorithms. 

\section*{Impact Statement}

Our contributions to sample-efficient MBRL have potential applications in robotics and autonomous systems, where data collection is costly, time-consuming, or safety-critical. By enabling more effective use of learned models, MRS.Q could reduce the real-world samples needed to train physical systems where extensive trial-and-error is impractical. While we evaluate only in simulation, the sample efficiency and performance gains demonstrated here may help accelerate the deployment of RL in real-world settings.

\bibliography{example_paper}

\begin{thebibliography}{76}
\providecommand{\natexlab}[1]{#1}
\providecommand{\url}[1]{\texttt{#1}}
\expandafter\ifx\csname urlstyle\endcsname\relax
  \providecommand{\doi}[1]{doi: #1}\else
  \providecommand{\doi}{doi: \begingroup \urlstyle{rm}\Url}\fi

\bibitem[Abbeel et~al.(2006)Abbeel, Quigley, and Ng]{abbeel2006using}
Abbeel, P., Quigley, M., and Ng, A.~Y.
\newblock Using inaccurate models in reinforcement learning.
\newblock In \emph{Proceedings of the 23rd international conference on Machine learning}, pp.\  1--8, 2006.

\bibitem[Amos et~al.(2021)Amos, Stanton, Yarats, and Wilson]{amos2021model}
Amos, B., Stanton, S., Yarats, D., and Wilson, A.~G.
\newblock On the model-based stochastic value gradient for continuous reinforcement learning.
\newblock In \emph{Learning for Dynamics and Control}, pp.\  6--20. PMLR, 2021.

\bibitem[An et~al.(2021)An, Moon, Kim, and Song]{an2021uncertainty}
An, G., Moon, S., Kim, J.-H., and Song, H.~O.
\newblock Uncertainty-based offline reinforcement learning with diversified q-ensemble.
\newblock \emph{Advances in neural information processing systems}, 34:\penalty0 7436--7447, 2021.

\bibitem[Asadi et~al.(2018)Asadi, Misra, and Littman]{asadi2018lipschitz}
Asadi, K., Misra, D., and Littman, M.
\newblock Lipschitz continuity in model-based reinforcement learning.
\newblock In \emph{International conference on machine learning}, pp.\  264--273. PMLR, 2018.

\bibitem[Atkeson \& Santamaria(1997)Atkeson and Santamaria]{atkeson1997comparison}
Atkeson, C.~G. and Santamaria, J.~C.
\newblock A comparison of direct and model-based reinforcement learning.
\newblock In \emph{Proceedings of international conference on robotics and automation}, volume~4, pp.\  3557--3564. IEEE, 1997.

\bibitem[Banijamali et~al.(2018)Banijamali, Shu, Bui, Ghodsi, et~al.]{banijamali2018robust}
Banijamali, E., Shu, R., Bui, H., Ghodsi, A., et~al.
\newblock Robust locally-linear controllable embedding.
\newblock In \emph{International Conference on Artificial Intelligence and Statistics}, pp.\  1751--1759. PMLR, 2018.

\bibitem[Brockman et~al.(2016)Brockman, Cheung, Pettersson, Schneider, Schulman, Tang, and Zaremba]{OpenAIGym}
Brockman, G., Cheung, V., Pettersson, L., Schneider, J., Schulman, J., Tang, J., and Zaremba, W.
\newblock Openai gym, 2016.

\bibitem[Buckman et~al.(2018)Buckman, Hafner, Tucker, Brevdo, and Lee]{buckman2018sample}
Buckman, J., Hafner, D., Tucker, G., Brevdo, E., and Lee, H.
\newblock Sample-efficient reinforcement learning with stochastic ensemble value expansion.
\newblock In \emph{Advances in Neural Information Processing Systems}, pp.\  8234--8244, 2018.

\bibitem[Chen et~al.(2020)Chen, Wang, Zhou, and Ross]{chen2020randomized}
Chen, X., Wang, C., Zhou, Z., and Ross, K.~W.
\newblock Randomized ensembled double q-learning: Learning fast without a model.
\newblock In \emph{International Conference on Learning Representations}, 2020.

\bibitem[Chua et~al.(2018)Chua, Calandra, McAllister, and Levine]{chua2018deep}
Chua, K., Calandra, R., McAllister, R., and Levine, S.
\newblock Deep reinforcement learning in a handful of trials using probabilistic dynamics models.
\newblock In \emph{Advances in Neural Information Processing Systems 31}, pp.\  4759--4770, 2018.

\bibitem[Clevert et~al.(2015)Clevert, Unterthiner, and Hochreiter]{clevert2015fast}
Clevert, D.-A., Unterthiner, T., and Hochreiter, S.
\newblock Fast and accurate deep network learning by exponential linear units (elus).
\newblock \emph{arXiv preprint arXiv:1511.07289}, 2015.

\bibitem[Deisenroth \& Rasmussen(2011)Deisenroth and Rasmussen]{deisenroth2011pilco}
Deisenroth, M. and Rasmussen, C.~E.
\newblock Pilco: A model-based and data-efficient approach to policy search.
\newblock In \emph{International Conference on Machine Learning}, pp.\  465--472, 2011.

\bibitem[Draeger et~al.(1995)Draeger, Engell, and Ranke]{draeger1995model}
Draeger, A., Engell, S., and Ranke, H.
\newblock Model predictive control using neural networks.
\newblock \emph{IEEE Control Systems Magazine}, 15\penalty0 (5):\penalty0 61--66, 1995.

\bibitem[Ebert et~al.(2017)Ebert, Finn, Lee, and Levine]{ebert2017self}
Ebert, F., Finn, C., Lee, A.~X., and Levine, S.
\newblock Self-supervised visual planning with temporal skip connections.
\newblock \emph{Conference on Robot Learning}, 2017.

\bibitem[Farebrother et~al.(2025)Farebrother, Pirotta, Tirinzoni, Munos, Lazaric, and Touati]{farebrother2025temporal}
Farebrother, J., Pirotta, M., Tirinzoni, A., Munos, R., Lazaric, A., and Touati, A.
\newblock Temporal difference flows.
\newblock In \emph{Forty-second International Conference on Machine Learning}, 2025.

\bibitem[Feinberg et~al.(2018)Feinberg, Wan, Stoica, Jordan, Gonzalez, and Levine]{feinberg2018model}
Feinberg, V., Wan, A., Stoica, I., Jordan, M.~I., Gonzalez, J.~E., and Levine, S.
\newblock Model-based value estimation for efficient model-free reinforcement learning.
\newblock \emph{arXiv preprint arXiv:1803.00101}, 2018.

\bibitem[Finn et~al.(2016)Finn, Tan, Duan, Darrell, Levine, and Abbeel]{finn2016deep}
Finn, C., Tan, X.~Y., Duan, Y., Darrell, T., Levine, S., and Abbeel, P.
\newblock Deep spatial autoencoders for visuomotor learning.
\newblock In \emph{2016 IEEE International Conference on Robotics and Automation (ICRA)}, pp.\  512--519. IEEE, 2016.

\bibitem[Fujimoto et~al.(2018)Fujimoto, van Hoof, and Meger]{fujimoto2018addressing}
Fujimoto, S., van Hoof, H., and Meger, D.
\newblock Addressing function approximation error in actor-critic methods.
\newblock In \emph{International Conference on Machine Learning}, volume~80, pp.\  1587--1596. PMLR, 2018.

\bibitem[Fujimoto et~al.(2019)Fujimoto, Meger, and Precup]{fujimoto2019off}
Fujimoto, S., Meger, D., and Precup, D.
\newblock Off-policy deep reinforcement learning without exploration.
\newblock In \emph{International Conference on Machine Learning}, pp.\  2052--2062, 2019.

\bibitem[Fujimoto et~al.(2020)Fujimoto, Meger, and Precup]{fujimoto2020equivalence}
Fujimoto, S., Meger, D., and Precup, D.
\newblock An equivalence between loss functions and non-uniform sampling in experience replay.
\newblock \emph{Advances in Neural Information Processing Systems}, 33, 2020.

\bibitem[Fujimoto et~al.(2024)Fujimoto, Chang, Smith, Gu, Precup, and Meger]{fujimoto2024sale}
Fujimoto, S., Chang, W.-D., Smith, E.~J., Gu, S.~S., Precup, D., and Meger, D.
\newblock For {SALE}: State-action representation learning for deep reinforcement learning.
\newblock In \emph{Thirty-seventh Conference on Neural Information Processing Systems}, 2024.

\bibitem[Fujimoto et~al.(2025)Fujimoto, D'Oro, Zhang, Tian, and Rabbat]{fujimoto2025towards}
Fujimoto, S., D'Oro, P., Zhang, A., Tian, Y., and Rabbat, M.
\newblock Towards general-purpose model-free reinforcement learning.
\newblock In \emph{The Thirteenth International Conference on Learning Representations}, 2025.

\bibitem[Gal et~al.(2016)Gal, McAllister, and Rasmussen]{gal2016improving}
Gal, Y., McAllister, R., and Rasmussen, C.~E.
\newblock Improving pilco with bayesian neural network dynamics models.
\newblock In \emph{Data-Efficient Machine Learning workshop, International Conference on Machine Learning}, 2016.

\bibitem[Glorot \& Bengio(2010)Glorot and Bengio]{glorot2010understanding}
Glorot, X. and Bengio, Y.
\newblock Understanding the difficulty of training deep feedforward neural networks.
\newblock In \emph{Proceedings of the thirteenth international conference on artificial intelligence and statistics}, pp.\  249--256. JMLR Workshop and Conference Proceedings, 2010.

\bibitem[Gu et~al.(2016)Gu, Lillicrap, Sutskever, and Levine]{gu2016continuous}
Gu, S., Lillicrap, T., Sutskever, I., and Levine, S.
\newblock Continuous deep q-learning with model-based acceleration.
\newblock In \emph{International conference on machine learning}, pp.\  2829--2838. PMLR, 2016.

\bibitem[Ha \& Schmidhuber(2018)Ha and Schmidhuber]{ha2018world}
Ha, D. and Schmidhuber, J.
\newblock Recurrent world models facilitate policy evolution.
\newblock \emph{Advances in neural information processing systems}, 31, 2018.

\bibitem[Hafner et~al.(2019)Hafner, Lillicrap, Fischer, Villegas, Ha, Lee, and Davidson]{hafner2019learning}
Hafner, D., Lillicrap, T., Fischer, I., Villegas, R., Ha, D., Lee, H., and Davidson, J.
\newblock Learning latent dynamics for planning from pixels.
\newblock In \emph{International conference on machine learning}, pp.\  2555--2565. PMLR, 2019.

\bibitem[Hafner et~al.(2023)Hafner, Pasukonis, Ba, and Lillicrap]{hafner2023mastering}
Hafner, D., Pasukonis, J., Ba, J., and Lillicrap, T.
\newblock Mastering diverse domains through world models.
\newblock \emph{arXiv preprint arXiv:2301.04104}, 2023.

\bibitem[Hansen et~al.(2024)Hansen, Su, and Wang]{hansen2024td}
Hansen, N., Su, H., and Wang, X.
\newblock Td-mpc2: Scalable, robust world models for continuous control.
\newblock In \emph{The Twelfth International Conference on Learning Representations}, 2024.

\bibitem[Hansen et~al.(2022)Hansen, Su, and Wang]{hansen2022temporal}
Hansen, N.~A., Su, H., and Wang, X.
\newblock Temporal difference learning for model predictive control.
\newblock In \emph{International Conference on Machine Learning}, pp.\  8387--8406. PMLR, 2022.

\bibitem[Henaff et~al.(2019)Henaff, Canziani, and LeCun]{henaff2019model}
Henaff, M., Canziani, A., and LeCun, Y.
\newblock Model-predictive policy learning with uncertainty regularization for driving in dense traffic.
\newblock In \emph{International Conference on Learning Representations}, 2019.

\bibitem[Higuera et~al.(2018)Higuera, Meger, and Dudek]{higuera2018synthesizing}
Higuera, J. C.~G., Meger, D., and Dudek, G.
\newblock Synthesizing neural network controllers with probabilistic model-based reinforcement learning.
\newblock In \emph{2018 IEEE/RSJ International Conference on Intelligent Robots and Systems (IROS)}, pp.\  2538--2544, 2018.

\bibitem[Janner et~al.(2019)Janner, Fu, Zhang, and Levine]{janner2019trust}
Janner, M., Fu, J., Zhang, M., and Levine, S.
\newblock When to trust your model: Model-based policy optimization.
\newblock \emph{Advances in Neural Information Processing Systems}, 32, 2019.

\bibitem[Janner et~al.(2022)Janner, Du, Tenenbaum, and Levine]{janner2022planning}
Janner, M., Du, Y., Tenenbaum, J., and Levine, S.
\newblock Planning with diffusion for flexible behavior synthesis.
\newblock In \emph{International Conference on Machine Learning}, pp.\  9902--9915. PMLR, 2022.

\bibitem[Lambert et~al.(2020)Lambert, Amos, Yadan, and Calandra]{lambert2020objective}
Lambert, N., Amos, B., Yadan, O., and Calandra, R.
\newblock Objective mismatch in model-based reinforcement learning.
\newblock In \emph{Learning for Dynamics and Control}, pp.\  761--770. PMLR, 2020.

\bibitem[Lambert et~al.(2021)Lambert, Wilcox, Zhang, Pister, and Calandra]{lambert2021learning}
Lambert, N., Wilcox, A., Zhang, H., Pister, K.~S., and Calandra, R.
\newblock Learning accurate long-term dynamics for model-based reinforcement learning.
\newblock In \emph{2021 60th IEEE Conference on decision and control (CDC)}, pp.\  2880--2887. IEEE, 2021.

\bibitem[Lambert et~al.(2022)Lambert, Pister, and Calandra]{lambert2022investigating}
Lambert, N., Pister, K., and Calandra, R.
\newblock Investigating compounding prediction errors in learned dynamics models.
\newblock \emph{arXiv preprint arXiv:2203.09637}, 2022.

\bibitem[Lan et~al.(2020)Lan, Pan, Fyshe, and White]{lan2020maxmin}
Lan, Q., Pan, Y., Fyshe, A., and White, M.
\newblock Maxmin q-learning: Controlling the estimation bias of q-learning.
\newblock In \emph{International Conference on Learning Representations}, 2020.

\bibitem[Lavoie et~al.(2023)Lavoie, Tsirigotis, Schwarzer, Vani, Noukhovitch, Kawaguchi, and Courville]{lavoie2023simplicial}
Lavoie, S., Tsirigotis, C., Schwarzer, M., Vani, A., Noukhovitch, M., Kawaguchi, K., and Courville, A.
\newblock Simplicial embeddings in self-supervised learning and downstream classification.
\newblock In \emph{The Eleventh International Conference on Learning Representations}, 2023.

\bibitem[Lee et~al.(2025)Lee, Lee, Seno, Kim, Stone, and Choo]{lee2025hyperspherical}
Lee, H., Lee, Y., Seno, T., Kim, D., Stone, P., and Choo, J.
\newblock Hyperspherical normalization for scalable deep reinforcement learning.
\newblock In \emph{Forty-second International Conference on Machine Learning}, 2025.

\bibitem[Levine \& Koltun(2013)Levine and Koltun]{levine2013guided}
Levine, S. and Koltun, V.
\newblock Guided policy search.
\newblock In \emph{International conference on machine learning}, pp.\  1--9. PMLR, 2013.

\bibitem[Lillicrap et~al.(2015)Lillicrap, Hunt, Pritzel, Heess, Erez, Tassa, Silver, and Wierstra]{DDPG}
Lillicrap, T.~P., Hunt, J.~J., Pritzel, A., Heess, N., Erez, T., Tassa, Y., Silver, D., and Wierstra, D.
\newblock Continuous control with deep reinforcement learning.
\newblock \emph{arXiv preprint arXiv:1509.02971}, 2015.

\bibitem[Lin et~al.(2025)Lin, Wang, Schneider, and Shi]{lin2025td}
Lin, H., Wang, P., Schneider, J., and Shi, G.
\newblock Td-m(pc)$^2$: Improving temporal difference mpc through policy constraint.
\newblock \emph{arXiv preprint arXiv:2502.03550}, 2025.

\bibitem[Loshchilov \& Hutter(2019)Loshchilov and Hutter]{loshchilov2018decoupled}
Loshchilov, I. and Hutter, F.
\newblock Decoupled weight decay regularization.
\newblock In \emph{International Conference on Learning Representations}, 2019.

\bibitem[Lowrey et~al.(2019)Lowrey, Rajeswaran, Kakade, Todorov, and Mordatch]{lowrey2019plan}
Lowrey, K., Rajeswaran, A., Kakade, S., Todorov, E., and Mordatch, I.
\newblock Plan online, learn offline: Efficient learning and exploration via model-based control.
\newblock In \emph{International Conference on Learning Representations}, 2019.

\bibitem[Ma et~al.(2024)Ma, Ni, Gehring, D’Oro, and Bacon]{ma2024transformer}
Ma, M., Ni, T., Gehring, C., D’Oro, P., and Bacon, P.-L.
\newblock Do transformer world models give better policy gradients?
\newblock In \emph{International Conference on Machine Learning}, pp.\  33855--33879. PMLR, 2024.

\bibitem[Mnih et~al.(2015)Mnih, Kavukcuoglu, Silver, Rusu, Veness, Bellemare, Graves, Riedmiller, Fidjeland, Ostrovski, et~al.]{DQN}
Mnih, V., Kavukcuoglu, K., Silver, D., Rusu, A.~A., Veness, J., Bellemare, M.~G., Graves, A., Riedmiller, M., Fidjeland, A.~K., Ostrovski, G., et~al.
\newblock Human-level control through deep reinforcement learning.
\newblock \emph{Nature}, 518\penalty0 (7540):\penalty0 529--533, 2015.

\bibitem[Mordatch \& Todorov(2014)Mordatch and Todorov]{mordatch2014combining}
Mordatch, I. and Todorov, E.
\newblock Combining the benefits of function approximation and trajectory optimization.
\newblock In \emph{Robotics: Science and Systems}, volume~4, pp.\ ~23, 2014.

\bibitem[Nagabandi et~al.(2018)Nagabandi, Kahn, Fearing, and Levine]{nagabandi2018neural}
Nagabandi, A., Kahn, G., Fearing, R.~S., and Levine, S.
\newblock Neural network dynamics for model-based deep reinforcement learning with model-free fine-tuning.
\newblock In \emph{2018 IEEE international conference on robotics and automation (ICRA)}, pp.\  7559--7566. IEEE, 2018.

\bibitem[Obando-Ceron et~al.(2026)Obando-Ceron, Mayor, Lavoie, Fujimoto, Courville, and Castro]{obando2025simplicial}
Obando-Ceron, J., Mayor, W., Lavoie, S., Fujimoto, S., Courville, A., and Castro, P.~S.
\newblock Simplicial embeddings improve sample efficiency in actor{\textendash}critic agents.
\newblock In \emph{The Fourteenth International Conference on Learning Representations}, 2026.

\bibitem[Oh et~al.(2015)Oh, Guo, Lee, Lewis, and Singh]{oh2015action}
Oh, J., Guo, X., Lee, H., Lewis, R.~L., and Singh, S.
\newblock Action-conditional video prediction using deep networks in atari games.
\newblock \emph{Advances in neural information processing systems}, 28, 2015.

\bibitem[Oh et~al.(2017)Oh, Singh, and Lee]{oh2017value}
Oh, J., Singh, S., and Lee, H.
\newblock Value prediction network.
\newblock \emph{Advances in neural information processing systems}, 30, 2017.

\bibitem[Palenicek et~al.(2023)Palenicek, Lutter, Carvalho, and Peters]{palenicek2023diminishing}
Palenicek, D., Lutter, M., Carvalho, J., and Peters, J.
\newblock Diminishing return of value expansion methods in model-based reinforcement learning.
\newblock In \emph{The Eleventh International Conference on Learning Representations}, 2023.

\bibitem[Paszke et~al.(2019)Paszke, Gross, Massa, Lerer, Bradbury, Chanan, Killeen, Lin, Gimelshein, Antiga, et~al.]{paszke2019pytorch}
Paszke, A., Gross, S., Massa, F., Lerer, A., Bradbury, J., Chanan, G., Killeen, T., Lin, Z., Gimelshein, N., Antiga, L., et~al.
\newblock Pytorch: An imperative style, high-performance deep learning library.
\newblock In \emph{Advances in Neural Information Processing Systems}, pp.\  8024--8035, 2019.

\bibitem[Schrittwieser et~al.(2020)Schrittwieser, Antonoglou, Hubert, Simonyan, Sifre, Schmitt, Guez, Lockhart, Hassabis, Graepel, et~al.]{schrittwieser2020mastering}
Schrittwieser, J., Antonoglou, I., Hubert, T., Simonyan, K., Sifre, L., Schmitt, S., Guez, A., Lockhart, E., Hassabis, D., Graepel, T., et~al.
\newblock Mastering atari, go, chess and shogi by planning with a learned model.
\newblock \emph{Nature}, 588\penalty0 (7839):\penalty0 604--609, 2020.

\bibitem[Schrittwieser et~al.(2021)Schrittwieser, Hubert, Mandhane, Barekatain, Antonoglou, and Silver]{schrittwieser2021online}
Schrittwieser, J., Hubert, T., Mandhane, A., Barekatain, M., Antonoglou, I., and Silver, D.
\newblock Online and offline reinforcement learning by planning with a learned model.
\newblock \emph{Advances in Neural Information Processing Systems}, 34:\penalty0 27580--27591, 2021.

\bibitem[Schwarzer et~al.(2023)Schwarzer, Ceron, Courville, Bellemare, Agarwal, and Castro]{schwarzer2023bigger}
Schwarzer, M., Ceron, J. S.~O., Courville, A., Bellemare, M.~G., Agarwal, R., and Castro, P.~S.
\newblock Bigger, better, faster: Human-level atari with human-level efficiency.
\newblock In \emph{International Conference on Machine Learning}, pp.\  30365--30380. PMLR, 2023.

\bibitem[Sferrazza et~al.(2024)Sferrazza, Huang, Lin, Lee, and Abbeel]{sferrazza2024humanoidbench}
Sferrazza, C., Huang, D.-M., Lin, X., Lee, Y., and Abbeel, P.
\newblock {HumanoidBench: Simulated Humanoid Benchmark for Whole-Body Locomotion and Manipulation}.
\newblock In \emph{Proceedings of Robotics: Science and Systems}, 2024.

\bibitem[Silver et~al.(2017)Silver, Hasselt, Hessel, Schaul, Guez, Harley, Dulac-Arnold, Reichert, Rabinowitz, Barreto, et~al.]{silver2017predictron}
Silver, D., Hasselt, H., Hessel, M., Schaul, T., Guez, A., Harley, T., Dulac-Arnold, G., Reichert, D., Rabinowitz, N., Barreto, A., et~al.
\newblock The predictron: End-to-end learning and planning.
\newblock In \emph{International Conference on Machine Learning}, pp.\  3191--3199. PMLR, 2017.

\bibitem[Sutton(1991)]{sutton1991dyna}
Sutton, R.~S.
\newblock Dyna, an integrated architecture for learning, planning, and reacting.
\newblock \emph{ACM Sigart Bulletin}, 2\penalty0 (4):\penalty0 160--163, 1991.

\bibitem[Sutton \& Barto(1998)Sutton and Barto]{suttonbarto}
Sutton, R.~S. and Barto, A.~G.
\newblock \emph{Reinforcement Learning: An Introduction}, volume~1.
\newblock MIT press Cambridge, 1998.

\bibitem[Talvitie(2014)]{talvitie2014model}
Talvitie, E.
\newblock Model regularization for stable sample rollouts.
\newblock In \emph{UAI}, pp.\  780--789, 2014.

\bibitem[Tassa et~al.(2018)Tassa, Doron, Muldal, Erez, Li, Casas, Budden, Abdolmaleki, Merel, Lefrancq, et~al.]{tassa2018deepmind}
Tassa, Y., Doron, Y., Muldal, A., Erez, T., Li, Y., Casas, D. d.~L., Budden, D., Abdolmaleki, A., Merel, J., Lefrancq, A., et~al.
\newblock Deepmind control suite.
\newblock \emph{arXiv preprint arXiv:1801.00690}, 2018.

\bibitem[Thrun \& Schwartz(1993)Thrun and Schwartz]{thrun1993bias}
Thrun, S. and Schwartz, A.
\newblock Issues in using function approximation for reinforcement learning.
\newblock In \emph{Proceedings of the 1993 Connectionist Models Summer School Hillsdale, NJ. Lawrence Erlbaum}, 1993.

\bibitem[Todorov et~al.(2012)Todorov, Erez, and Tassa]{mujoco}
Todorov, E., Erez, T., and Tassa, Y.
\newblock Mujoco: A physics engine for model-based control.
\newblock In \emph{IEEE/RSJ International Conference on Intelligent Robots and Systems (IROS)}, pp.\  5026--5033. IEEE, 2012.

\bibitem[Towers et~al.(2024)Towers, Kwiatkowski, Terry, Balis, De~Cola, Deleu, Goul{\~a}o, Kallinteris, Krimmel, KG, et~al.]{towers2024gymnasium}
Towers, M., Kwiatkowski, A., Terry, J., Balis, J.~U., De~Cola, G., Deleu, T., Goul{\~a}o, M., Kallinteris, A., Krimmel, M., KG, A., et~al.
\newblock Gymnasium: A standard interface for reinforcement learning environments.
\newblock \emph{arXiv preprint arXiv:2407.17032}, 2024.

\bibitem[Van~Hasselt(2010)]{hasselt2010double}
Van~Hasselt, H.
\newblock Double q-learning.
\newblock In \emph{Advances in Neural Information Processing Systems}, pp.\  2613--2621, 2010.

\bibitem[Van~Hasselt et~al.(2016)Van~Hasselt, Guez, and Silver]{DoubleDQN}
Van~Hasselt, H., Guez, A., and Silver, D.
\newblock Deep reinforcement learning with double q-learning.
\newblock In \emph{AAAI}, pp.\  2094--2100, 2016.

\bibitem[Van~Hasselt et~al.(2019)Van~Hasselt, Hessel, and Aslanides]{van2019use}
Van~Hasselt, H.~P., Hessel, M., and Aslanides, J.
\newblock When to use parametric models in reinforcement learning?
\newblock \emph{Advances in Neural Information Processing Systems}, 32, 2019.

\bibitem[Venkatraman et~al.(2015)Venkatraman, Hebert, and Bagnell]{venkatraman2015improving}
Venkatraman, A., Hebert, M., and Bagnell, J.
\newblock Improving multi-step prediction of learned time series models.
\newblock In \emph{Proceedings of the AAAI Conference on Artificial Intelligence}, volume~29, 2015.

\bibitem[Wang et~al.(2024)Wang, Liu, Ye, You, and Gao]{wang2024efficientzero}
Wang, S., Liu, S., Ye, W., You, J., and Gao, Y.
\newblock Efficientzero v2: Mastering discrete and continuous control with limited data.
\newblock In \emph{International Conference on Machine Learning}, pp.\  51041--51062. PMLR, 2024.

\bibitem[Wang et~al.(2025)Wang, Guo, Wang, Qian, and Lan]{wang2025bootstrapped}
Wang, Y., Guo, H., Wang, S., Qian, L., and Lan, X.
\newblock Bootstrapped model predictive control.
\newblock In \emph{The Thirteenth International Conference on Learning Representations}, 2025.

\bibitem[Watter et~al.(2015)Watter, Springenberg, Boedecker, and Riedmiller]{watter2015embed}
Watter, M., Springenberg, J., Boedecker, J., and Riedmiller, M.
\newblock Embed to control: A locally linear latent dynamics model for control from raw images.
\newblock \emph{Advances in neural information processing systems}, 28, 2015.

\bibitem[Williams et~al.(2015)Williams, Aldrich, and Theodorou]{williams2015model}
Williams, G., Aldrich, A., and Theodorou, E.
\newblock Model predictive path integral control using covariance variable importance sampling.
\newblock \emph{arXiv preprint arXiv:1509.01149}, 2015.

\bibitem[Ye et~al.(2021)Ye, Liu, Kurutach, Abbeel, and Gao]{ye2021mastering}
Ye, W., Liu, S., Kurutach, T., Abbeel, P., and Gao, Y.
\newblock Mastering atari games with limited data.
\newblock \emph{Advances in neural information processing systems}, 34:\penalty0 25476--25488, 2021.

\bibitem[Zhan et~al.(2025)Zhan, Wang, Zhang, Gao, Tomizuka, and Li]{zhan2025bootstrap}
Zhan, G., Wang, L., Zhang, X., Gao, J., Tomizuka, M., and Li, S.~E.
\newblock Bootstrap off-policy with world model.
\newblock In \emph{The Thirty-ninth Annual Conference on Neural Information Processing Systems}, 2025.

\end{thebibliography}
\bibliographystyle{icml2026}

\newpage
\appendix
\onecolumn

\renewcommand \thepart{}
\renewcommand \partname{}

\part{Appendix} \label{appendix}

\crefname{appendix}{Appendix}{Appendices}
\Crefname{appendix}{Appendix}{Appendices}
\crefalias{section}{appendix}
\crefalias{subsection}{appendix}

\startcontents[appendix]

\vspace{-20pt}
\noindent\rule{\textwidth}{0.5pt}
\vspace{-20pt}

{\small

\printcontents[appendix]{l}{1}

}
\noindent\rule{\textwidth}{0.5pt}

\section{Experimental Details}

\subsection{Hyperparameters} 

\begin{table}[ht]
\vspace{-64pt}
\caption{\textbf{MRS.Q Hyperparameters.} \hlfancy{sb_blue!25}{New or changed} hyperparameters from MR.Q are highlighted. Hyperparameters values are kept fixed across all benchmarks. MPC hyperparameters come directly from TD-MPC2~\citep{hansen2024td}, except for the policy standard deviation.}\label{table:hyperparameters}
\centering
\small
\begin{tabular}{cll}
\toprule
& Hyperparameter & Value \\
\midrule
\multirow{6}{*}{\shortstack{MR.Q\\\citep{fujimoto2025towards}}}  & Dynamics loss weight $\lambda_\text{Dynamics}$ & \cellcolor{sb_blue!25}{$1 \rightarrow 20$} \\
& Reward loss weight $\lambda_\text{Reward}$ & $0.1$ \\
& Terminal loss weight $\lambda_\text{Terminal}$ & \cellcolor{sb_blue!25}{$0.1 \rightarrow 1$} \\
& Pre-activation loss weight $\lambda_\text{pre-activ}$ & $1\text{e}-5$ \\
& Encoder horizon $H_\text{Enc}$ & $5$ \\
& Multi-step returns horizon $H_Q$ & $3$ \\ 
\midrule
\multirow{2}{*}{\shortstack{TD3\\\citep{fujimoto2018addressing}}} 
& Target policy noise~$\sigma$      & $\N(0,0.2^2)$ \\
& Target policy noise clipping~$c$  & $(-0.3, 0.3)$ \\
\midrule
\multirow{2}{*}{\shortstack{LAP\\\citep{fujimoto2020equivalence}}} 
& Probability smoothing $\alpha$    & $0.4$ \\
& Minimum priority                  & $1$ \\
\midrule
\multirow{2}{*}{Exploration} & Initial random exploration time steps & $10$k \\
& Exploration noise    & \cellcolor{sb_blue!25}$\N(0,0.2^2) \rightarrow \N(0,0)$ \\
\midrule
\multirow{5}{*}{Common}
& Discount factor~$\y$      & $0.99$ \\
& Replay buffer capacity    & $1$M \\
& Mini-batch size           & $256$ \\
& Target update frequency~$T_\text{target}$   & $250$ \\
& Replay ratio              & $1$ \\ 
\midrule
\multirow{13}{*}{Encoder Network} 
& Optimizer        & AdamW~\citep{loshchilov2018decoupled} \\
& Learning rate    & $1\text{e}-4$ \\
& Weight decay     & $1\text{e}-4$ \\
& $\mathbf{z}_s$ dim & $512$ \\
& $\mathbf{z}_{sa}$ dim & $512$ \\
& $\mathbf{z}_a$ dim (only used within architecture) & $256$ \\
& Hidden dim & $512$ \\
& Activation function & ELU~\citep{clevert2015fast} \\
& Weight initialization & Xavier uniform~\citep{glorot2010understanding} \\
& Bias initialization & $0$ \\
& Reward bins & $65$ \\
& Reward range & $[-10,10]$ (effective: $[-22\text{k}, 22\text{k}]$) \\
& SEM Groups & \cellcolor{sb_blue!25}8 \\
\midrule
\multirow{8}{*}{Value Network} 
& Optimizer        & AdamW \\
& Learning rate    & $3\text{e}-4$ \\
& Hidden dim & $512$ \\
& Activation function & ELU \\
& Weight initialization & Xavier uniform \\
& Bias initialization & $0$ \\
& Gradient clip norm & $20$ \\
& Ensemble size & \cellcolor{sb_blue!25}$2 \rightarrow 10$ \\
\midrule
\multirow{6}{*}{Policy Network} 
& Optimizer        & AdamW \\
& Learning rate    & $3\text{e}-4$ \\
& Hidden dim & $512$ \\
& Activation function & ReLU \\
& Weight initialization & Xavier uniform \\
& Bias initialization & $0$ \\ 
\midrule
\multirow{8}{*}{\shortstack{\hlfancy{sb_blue!25}{MPC}\\\citep{williams2015model}\\\citep{hansen2024td}}}
& MPC horizon $H$ & 3 \\
& Number of MPC iterations $I$ & 6 \\ 
& Number of samples $n$ & 512 \\
& Number of policy actions $n_\pi$ & 24 \\
& Number of elites $k$ & 64 \\ 
& Policy standard deviation $\sigma_\text{det}$ & \cellcolor{sb_blue!25}0.1 \\
& Max standard deviation $\sigma_\text{max}$ & 2 \\
& Min standard deviation $\sigma_\text{min}$ & 0.05 \\
& Temperature $\tau$ & 0.5 \\ 
\bottomrule
\end{tabular}
\vspace{-48pt}
\end{table}

\clearpage

\subsection{MPC Procedure}

For our search algorithm, we follow the TD-MPC2~\citep{hansen2024td} version of Model Predictive Path Integral (MPPI) control~\citep{williams2015model}. Let $f(s) = \mathbf{z}_s$ be the state encoder, let $g(\mathbf{z}_s, a) = \mathbf{z}_{sa}$ be the state-action encoder, and let $h(\mathbf{z}_{sa})$ be the dynamics prediction model. Let $t_\text{episode}$ denote the current time step of the episode. 

\begin{figure}[ht]
\phantomcaption \label{alg:mpc}
\centering
\begin{NiceTabular}{l >{\color{gray}}r}
\toprule
\textbf{Algorithm 1} \texttt{MPC} \\
\midrule
\textbf{Input} \\
\hspace{2pt}\textbullet\hspace{5pt}State $s$ \\
\textbf{Hyperparameters} \\
\hspace{2pt}\textbullet\hspace{5pt}MPC horizon $H$ \\
\hspace{2pt}\textbullet\hspace{5pt}Number of MPC iterations $I$ \\ 
\hspace{2pt}\textbullet\hspace{5pt}Number of samples $n$ \\
\hspace{2pt}\textbullet\hspace{5pt}Number of policy actions $n_\pi$ \\
\hspace{2pt}\textbullet\hspace{5pt}Number of elites $k$ \\
\hspace{2pt}\textbullet\hspace{5pt}Policy standard deviation $\sigma_\text{det}$ \\
\hspace{2pt}\textbullet\hspace{5pt}Max standard deviation $\sigma_\text{max}$ \\
\hspace{2pt}\textbullet\hspace{5pt}Min standard deviation $\sigma_\text{min}$ \\
\hspace{2pt}\textbullet\hspace{5pt}Temperature $\tau$ \\ 
\midrule
\textcolor{gray}{Sample policy actions} \\
$\mathbf{z}_s = f(s)$ & Encode initial state \\
\textbf{For} $t=0$ \textbf{to} $H-1$ \textbf{do}: \\
\qquad $a^\pi_t = \{ \N(\pi(\mathbf{z}_s), \sigma_\text{det}) \}_{n_\pi} $ & $a^\pi_t$ is a set of $n_\pi$ actions \\
\qquad $\mathbf{z}_s = h(g(\mathbf{z}_s, a^\pi_t$)) & Roll the model forward \\
\midrule
\multicolumn{2}{l}{\textcolor{gray}{Initialize mean and standard deviation of randomly sampled actions}} \\
$\mu_t = \mu^\text{prev}_{t+1}$ \textbf{if} $t_\text{episode} > 0$ \textbf{else} $\mu_t = 0$ & Initialize random action mean using results from a prior MPPI call \\
$\sigma_t = \sigma_\text{max}$ & Initialize standard deviation \\
\midrule 
\textcolor{gray}{MPC iterations} \\
\textbf{For} $i=0$ \textbf{to} $I-1$ \textbf{do}: \\
\qquad $a^\text{rand}_t = \{ N(\mu_t, \sigma_t) \}_{n - n_\pi} $ & $a^\text{rand}_t$ is a set of $n - n_\pi$ actions \\
\qquad $a_t = [a^\pi_t, a^\text{rand}_t]$ & Actions are both the policy actions and the sampled actions \\
\qquad $V = \texttt{EstimateValue}(s, a_t)$ & Compute the value of each action sequence $\{a_0, a_1, ... \}$ \\
\qquad $V_\text{elite}, a_\text{elite} = \texttt{Top-k}(V, k)$ & Select the highest $k$ valued actions \\
\qquad $s = e^{\tau (V_\text{elite} - \max V_\text{elite})}$ & Score of each action, using their value and temperature $\tau$ \\
\qquad $s \leftarrow \frac{s}{\sum s}$ & Normalize scores \\ 
\qquad $\mu = \sum s a_\text{elite}$ & Compute new mean \\
\qquad $\sigma = \sqrt{\sum s (a_\text{elite} - \mu)^2}$ & Compute new standard deviation \\ 
\qquad $\sigma = \text{clamp}(\sigma, \sigma_\text{min}, \sigma_\text{max})$ \\
$\mu_\text{prev} = \mu$ & Save mean for \texttt{MPC} call at the next episode time step \\
\textbf{Return} $\texttt{GumbelSoftMaxSample}(a_\text{elite}, s, \tau)$ & Sample final action based on scores \\
\bottomrule
\end{NiceTabular}
\end{figure}

\begin{figure}[ht]
\phantomcaption \label{alg:value}
\centering
\begin{NiceTabular}{l >{\color{gray}}r}
\toprule
\textbf{Algorithm 2} \texttt{EstimateValue} \\
\midrule
\textbf{Input} \\
\hspace{2pt}\textbullet\hspace{5pt}State $s$ \\
\hspace{2pt}\textbullet\hspace{5pt}Sequence of actions $a_t$ \\
\textbf{Hyperparameters} \\
\hspace{2pt}\textbullet\hspace{5pt}MPC horizon $H$ \\
\hspace{2pt}\textbullet\hspace{5pt}Discount $\y$ \\
\midrule
$R = 0, d = 1, g = 1$ & Initialize tracking variables \\
$\mathbf{z}_s = f(s)$ & Encode initial state \\
\textbf{For} $t=0$ \textbf{to} $H-1$ \textbf{do}: \\
\qquad $\mathbf{z}_s, \tilde{r}, \tilde{d} = h(g(\mathbf{z}_s, a_t$)) & Roll the model forward, predicting next embedding $\mathbf{z}_s$, reward $\tilde{r}$, and termination $\tilde{d}$ \\
\qquad $R = R + g d \tilde{r}$ & Update accumulated return \\ 
\qquad $d = 1 - \texttt{round}(\tilde{d})$ & Update accumulated termination, rounding to make predicted termination $\in \{0,1\}$ \\ 
\qquad $g = \y g$ & Update accumulated discount \\
$a_H = \pi(\mathbf{z}_s)$ & Final policy action \\
$Q = \min_i Q_i(g(\mathbf{z}_s, a_H))$ & Final value, taking minimum over the ensemble \\
\textbf{Return} $R + g d Q$ & Compute and return the final value of the action sequence \\
\bottomrule
\end{NiceTabular}
\end{figure}

\clearpage

\subsection{Network Architecture}

This section describes MRS.Q's network architecture using PyTorch code blocks~\citep{paszke2019pytorch}. The network definitions follow MR.Q's architecture, with minor modifications to incorporate SEM~\citep{lavoie2023simplicial}.

\textbf{Preamble}
\begin{lstlisting}
import torch
import torch.nn as nn
import torch.nn.functional as F

zs_dim = 512
za_dim = 256
zsa_dim = 512

def ln(x):
    return F.layer_norm(x, (x.shape[-1],))

def SEM(self, x):
    shape = x.shape
    x = x.reshape(*shape[:-1], -1, 8)
    x = F.softmax(x, dim=-1)
    return x.reshape(*shape)
\end{lstlisting}

\textbf{State Encoder $f$ Network}

The state encoder is a three-layer MLP with a hidden dimension of 512. Each hidden layer is followed by LayerNorm and an ELU activation. The final layer applies LayerNorm with a learnable affine transformation, followed by SEM.

The resulting state embedding $\mathbf{z}_s$ is trained end-to-end with the state-action encoder.

\begin{lstlisting}
self.zs1 = nn.Linear(state_dim, 512)
self.zs2 = nn.Linear(512, 512)
self.zs3 = nn.Linear(512, zs_dim)
self.ln = nn.LayerNorm(zs_dim)

self.activ = F.elu

def forward(self, state):
    zs = self.activ(ln(self.zs1(state)))
    zs = self.activ(ln(self.zs2(zs)))
    return SEM(self.ln(self.zs3(zs)))
\end{lstlisting}

\textbf{State-Action Encoder $g$ Network}

The action input passes through a linear layer followed by an ELU activation. The processed action is then concatenated with the state embedding and fed through a three-layer MLP with a hidden dimension of 512. LayerNorm and ELU activations follow the first two layers.

The resulting state-action embedding $\mathbf{z}_{sa}$ is passed to a linear layer that predicts the next state embedding, reward, and terminal signal. The final layer of the dynamics model applies LayerNorm with a learnable affine transformation, followed by SEM.

\begin{lstlisting}
self.za = nn.Linear(action_dim, za_dim)
self.zsa1 = nn.Linear(zs_dim + za_dim, 512)
self.zsa2 = nn.Linear(512, 512)
self.zsa3 = nn.Linear(512, zsa_dim)

self.ln = nn.LayerNorm(zs_dim)
self.dynamics_model = nn.Linear(zsa_dim, zs_dim)
self.reward_model = nn.Linear(zsa_dim, 65)
self.terminal_model = nn.Linear(zsa_dim, 1)

self.activ = F.elu

def forward(self, zs, action):
    za = self.activ(self.za(action))
    zsa = torch.cat([zs, za], 1)
    zsa = self.activ(ln(self.zsa1(zsa)))
    zsa = self.activ(ln(self.zsa2(zsa)))
    zsa = self.zsa3(zsa)
    return (
        SEM(self.ln(self.dynamics_model(zsa))), 
        self.reward_model(zsa), 
        self.terminal_model(zsa), 
        zsa
    )
\end{lstlisting}

\textbf{Value $Q$ Networks}

The value network is a four-layer MLP with a hidden dimension of 512. LayerNorm and ELU activations follow the first three layers. An ensemble of 10 value networks is used, each sharing the same architecture and forward pass.

\begin{lstlisting}
self.l1 = nn.Linear(zsa_dim, 512)
self.l2 = nn.Linear(512, 512)
self.l3 = nn.Linear(512, 512)
self.l4 = nn.Linear(512, 1)

self.activ = F.elu

def forward(self, zsa):
    q = self.activ(ln(self.l1(zsa)))
    q = self.activ(ln(self.l2(q)))
    q = self.activ(ln(self.l3(q)))
    return self.l4(q)
\end{lstlisting}

\textbf{Policy $\pi$ Network}

The policy network is a three-layer MLP with a hidden dimension of 512. LayerNorm and ReLU activations follow the first two layers, with a tanh function applied to the output.

\begin{lstlisting}
self.l1 = nn.Linear(zs_dim, 512)
self.l2 = nn.Linear(hdim, 512)
self.l3 = nn.Linear(512, action_dim)

self.activ = F.relu

def forward(self, zs):
    a = self.activ(ln(self.l1(zs)))
    a = self.activ(ln(self.l2(a)))
    return torch.tanh(self.l3(a))
\end{lstlisting}

\clearpage

\subsection{Environments} \label{appendix:envs}

All main experiments for MRS.Q and baseline methods used 10 random seeds, while ablation studies used 5 seeds. Evaluations were performed every 5k time steps, taking the average performance across 10 episodes.

\textbf{Gym}~\citep{mujoco, OpenAIGym, towers2024gymnasium}. We evaluate on the same five environments used by MR.Q~\citep{fujimoto2025towards}, all based on the \texttt{-v4} version with no preprocessing applied. Following MR.Q, we report aggregate scores using TD3-normalized performance, with TD3 reference scores obtained from TD7~\citep{fujimoto2024sale}:
\begin{equation}
    \text{TD3-Normalized}(x) := \frac{x - \text{random score}}{\text{TD3 score} - \text{random score}}.
\end{equation}

\begin{table}[ht]
\small
\centering
\caption{Scores used to normalize Gym tasks.}\label{tab:td3-normalized}
\begin{tabular}{lrr}
\toprule
& Random & TD3 \\
\midrule
Ant & -70.288 & \z3942\\
HalfCheetah & -289.415 & 10574\\
Hopper & 18.791 & \z3226\\
Humanoid & 120.423 & \z5165 \\
Walker2d & 2.791 & \z3946 \\
\bottomrule
\end{tabular}
\end{table}

\textbf{DM Control Suite (DMC)}~\citep{tassa2018deepmind}. We evaluate on the same 28 environments used by MR.Q, with an action repeat of 2 following prior work~\citep{hansen2024td, fujimoto2025towards}.

\textbf{HumanoidBench} \citep{sferrazza2024humanoidbench}. We evaluate on the same set of 14 \texttt{-v0} environments for both HumanoidBench experiment configurations. Following prior work~\citep{wang2025bootstrapped}, we omit the reach task from aggregate scores due to its different reward scale. Experiments without hands use the \texttt{h1} environments (e.g., \texttt{h1-walk-v0}), while experiments with hands use the \texttt{h1-hand} environments (e.g., \texttt{h1hand-walk-v0}). For action repeat settings, MRS.Q and the model-based baselines (TD-MPC2, BMPC, BOOM) use an action repeat of 1, following author-provided hyperparameters. The model-free baselines (MR.Q, SimbaV2) use an action repeat of 2; MR.Q+MPC also uses an action repeat of 2 to ensure a fair comparison with MR.Q.

\subsection{Baselines} \label{app:baselines}

All experiments were run for 10 seeds over 1M environment steps and use default author-provided hyperparameters for all tasks. 

\textbf{MR.Q}~\citep{fujimoto2025towards}. Results were obtained from the authors' GitHub repository (\url{https://github.com/facebookresearch/MRQ}), commit \textcolor{gray}{280d9c0263964463522d16ec2daee57b1b7bf087}, except for HumanoidBench, where we re-ran the authors' code. For HumanoidBench, we use an action repeat of 2, as this achieved a slightly higher performance than an action repeat of 1.

\textbf{TD-MPC2}~\citep{hansen2024td}. Results were obtained by re-running the authors' code (\url{https://github.com/nicklashansen/tdmpc2}, commit \textcolor{gray}{8bbc14ebabdb32ea7ada5c801dc525d0dc73bafe}). Unlike the MR.Q paper, which used separate TD-MPC2 codebases for terminal and non-terminal environments, we use this single codebase for all experiments, as it is designed to handle both settings.

\textbf{TD-M(PC)$^2$} \citep{lin2025td}. Results were obtained by re-running the authors' code (\url{https://github.com/DarthUtopian/tdmpc_square_public}), commit \textcolor{gray}{d1c2632c36effd2f7b661bfe5f822a3db8054d40}. To enable termination prediction for Gym tasks, we extend the dynamics model to include termination signals, following the TD-MPC2 implementation. This modification applies only to Gym tasks with termination conditions.

\textbf{BMPC} \citep{wang2025bootstrapped}. Results were obtained by re-running the authors' code (\url{https://github.com/wertyuilife2/bmpc}), commit \textcolor{gray}{6746ea7d265a8b82c8347fd7e894373ce00333fa}. To enable termination prediction for Gym tasks, we extend the dynamics model to include termination signals, following the TD-MPC2 implementation. This modification applies only to Gym tasks with termination conditions.

\textbf{BOOM} \citep{zhan2025bootstrap}. Results were obtained by re-running the authors' code (\url{https://github.com/molumitu/BOOM_MBRL}), commit \textcolor{gray}{9b3156d1fadda5ca5318274c404b694ad7b0786f}. To enable termination prediction for Gym tasks, we extend the dynamics model to include termination signals, following the TD-MPC2 implementation. This modification applies only to Gym tasks with termination conditions.

\textbf{SimbaV2} \citep{lee2025hyperspherical}. Results were obtained by re-running the authors' code (\url{https://github.com/DAVIAN-Robotics/SimbaV2}), commit \textcolor{gray}{86899c277cdc697b2b02d827243de1ea93f20a1d}. 

\subsection{Model and Value Accuracy Experiments}

To compute the error terms reported in \Cref{table:acc_perf} and \Cref{fig:value_error}, we evaluate error terms on state-action pairs sampled according to a fully trained agent. Specifically, for each algorithm, we use the final policy to collect 50 state-action pairs along a trajectory at regular intervals of 20 time steps, resetting the environment as needed.

We obtain ground-truth value estimates by resetting the internal simulator to each recorded state, executing the initial action, and then following the agent's policy until episode termination. We repeat this process 100 times per state-action pair to ensure reliable estimates.

\clearpage

\section{Computational Cost} \label{sec:appendix_cost}

We evaluate the computational cost of MRS.Q against our baselines using consistent hardware, with results reported in \Cref{appendix:table:cost}. For a fair comparison, \texttt{torch.compile} is disabled for all approaches. All experiments are conducted on HalfCheetah-v4 for 1M time steps.

\begin{table}[ht]
\centering
\small
\caption{Computational cost.} \label{appendix:table:cost}
\begin{tabular}{lll}
\toprule
Method                    & MPC & Wall-clock time \\ 
\midrule
MR.Q                      & \textcolor{sb_red}{$\times$} & 10 hours                     \\
MR.Q + 10 value functions & \textcolor{sb_red}{$\times$} & 16 hours                     \\
TD-MPC2                   & \textcolor{sb_green}{\checkmark} & 45 hours                     \\
MRS.Q                     & \textcolor{sb_green}{\checkmark} & 53 hours                     \\
\bottomrule
\end{tabular}
\end{table}

The results show that although MRS.Q incurs significantly higher computational cost than MR.Q, this overhead is primarily due to MPC rather than the ensemble of 10 value functions.

\clearpage

\section{Full Results}

\subsection{Gym}

\begin{table}[ht]
\centering
\tiny
\caption{\textbf{Gym results.} Average final performance at 1M time steps. \textcolor{gray}{[Bracketed values]} indicate 95\% stratified bootstrap confidence intervals over 10 seeds.}
\setlength{\tabcolsep}{0pt}
\begin{tabularx}{\textwidth}{l@{\hspace{4pt}} rX@{\hspace{4pt}} rX@{\hspace{4pt}} rX@{\hspace{4pt}} rX@{\hspace{4pt}} rX@{\hspace{4pt}} rX@{\hspace{4pt}} rX@{\hspace{4pt}} rX}
\toprule
Task & \multicolumn{2}{c}{MRS.Q} & \multicolumn{2}{c}{MR.Q} & \multicolumn{2}{c}{MR.Q+MPC} & \multicolumn{2}{c}{TD-MPC2} & \multicolumn{2}{c}{TD-M(PC)$^2$} & \multicolumn{2}{c}{BMPC} & \multicolumn{2}{c}{BOOM} & \multicolumn{2}{c}{SimbaV2} \\
\midrule
Ant & 7189 & ~\textcolor{gray}{[6394, 7804]} & 6901 & ~\textcolor{gray}{[6261, 7482]} & 7068 & ~\textcolor{gray}{[6627, 7865]} & 1938 & ~\textcolor{gray}{[795, 3694]} & 3516 & ~\textcolor{gray}{[1899, 5336]} & 5102 & ~\textcolor{gray}{[2664, 6966]} & 4740 & ~\textcolor{gray}{[3206, 6167]} & 6912 & ~\textcolor{gray}{[6545, 7192]} \\
HalfCheetah & 14649 & ~\textcolor{gray}{[14149, 15158]} & 12939 & ~\textcolor{gray}{[11663, 13762]} & 4544 & ~\textcolor{gray}{[1083, 9189]} & 11188 & ~\textcolor{gray}{[6871, 15475]} & 12419 & ~\textcolor{gray}{[10716, 13776]} & 14319 & ~\textcolor{gray}{[11251, 16002]} & 12289 & ~\textcolor{gray}{[8333, 16247]} & 13168 & ~\textcolor{gray}{[13039, 13281]} \\
Hopper & 3248 & ~\textcolor{gray}{[2945, 3516]} & 2692 & ~\textcolor{gray}{[2131, 3309]} & 702 & ~\textcolor{gray}{[141, 1313]} & 267 & ~\textcolor{gray}{[170, 360]} & 334 & ~\textcolor{gray}{[232, 439]} & 32 & ~\textcolor{gray}{[8, 64]} & 196 & ~\textcolor{gray}{[143, 253]} & 3483 & ~\textcolor{gray}{[2972, 3948]} \\
Humanoid & 10526 & ~\textcolor{gray}{[10186, 10816]} & 10223 & ~\textcolor{gray}{[9929, 10498]} & 4531 & ~\textcolor{gray}{[868, 9729]} & 297 & ~\textcolor{gray}{[260, 336]} & 3003 & ~\textcolor{gray}{[1724, 4239]} & 220 & ~\textcolor{gray}{[182, 258]} & 1949 & ~\textcolor{gray}{[1046, 3480]} & 9326 & ~\textcolor{gray}{[8213, 10339]} \\
\bottomrule
\end{tabularx}
\end{table}

\begin{figure}[ht]

\begin{tikzpicture}[trim axis right]
    \begin{axis}[
        height=0.1615\textwidth,
        width=0.19\textwidth,
        title={\vphantom{p}Ant},
        ylabel={Total Reward},
        xlabel={Time steps (1M)},
        xtick={0.0, 40.0, 80.0, 120.0, 160.0, 200.0},
        xticklabels={0.0, 0.2, 0.4, 0.6, 0.8, 1.0},
        ytick={-2.4539877300613497, 0.0, 2.4539877300613497, 4.9079754601226995, 7.361963190184049, 9.815950920245399, 12.269938650306749},
        yticklabels={\hphantom{00}-2k, \hphantom{-000}0, \hphantom{-00}2k, \hphantom{-00}4k, \hphantom{-00}6k, \hphantom{-00}8k, \hphantom{-0}10k},
    ]
    \addplot graphics [
    ymin=-1.3365834338930793, ymax=10.000904189470903,
    xmin=-10.0, xmax=210.0,
    ]{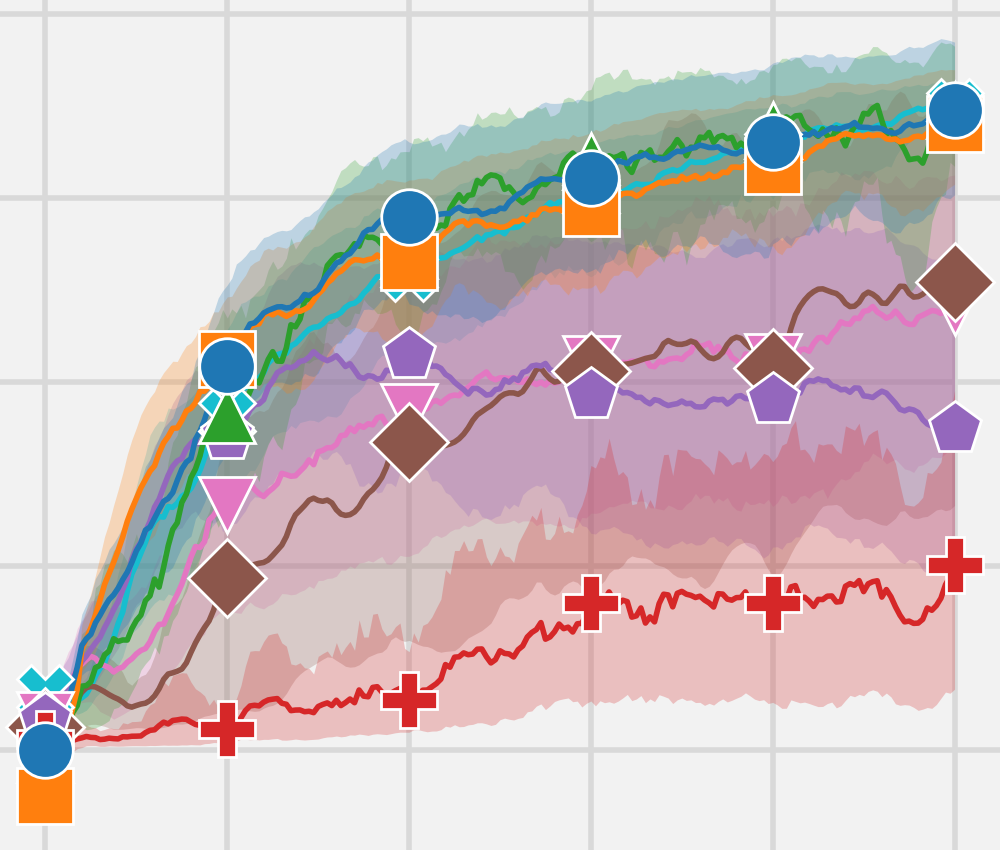};
    \end{axis}
\end{tikzpicture}
\begin{tikzpicture}[trim axis right]
    \begin{axis}[
        height=0.1615\textwidth,
        width=0.19\textwidth,
        title={\vphantom{p}HalfCheetah},
        ylabel={},
        xlabel={Time steps (1M)},
        xtick={0.0, 40.0, 80.0, 120.0, 160.0, 200.0},
        xticklabels={0.0, 0.2, 0.4, 0.6, 0.8, 1.0},
        ytick={-1.4602803738317758, 0.0, 1.4602803738317758, 2.9205607476635516, 4.380841121495327, 5.841121495327103, 7.3014018691588785, 8.761682242990654, 10.22196261682243},
        yticklabels={\hphantom{}-2.5k, \hphantom{-000}0, \hphantom{-}2.5k, \hphantom{-00}5k, \hphantom{-}7.5k, \hphantom{-0}10k, \hphantom{-}12.5k, \hphantom{-0}15k, \hphantom{-}17.5k},
    ]
    \addplot graphics [
    ymin=-0.6802088322986369, ymax=10.000284975817276,
    xmin=-10.0, xmax=210.0,
    ]{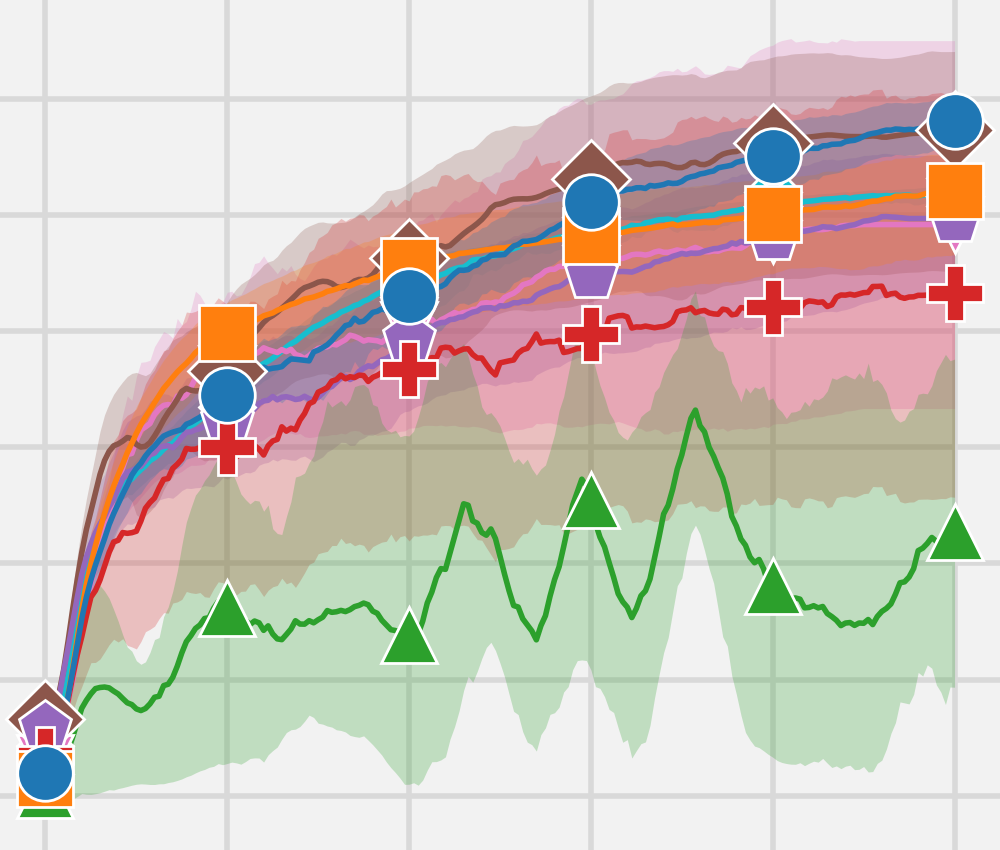};
    \end{axis}
\end{tikzpicture}
\begin{tikzpicture}[trim axis right]
    \begin{axis}[
        height=0.1615\textwidth,
        width=0.19\textwidth,
        title={\vphantom{p}Hopper},
        ylabel={},
        xlabel={Time steps (1M)},
        xtick={0.0, 40.0, 80.0, 120.0, 160.0, 200.0},
        xticklabels={0.0, 0.2, 0.4, 0.6, 0.8, 1.0},
        ytick={-1.1737089201877935, 0.0, 1.1737089201877935, 2.347417840375587, 3.5211267605633805, 4.694835680751174, 5.868544600938967, 7.042253521126761, 8.215962441314554, 9.389671361502348, 10.56338028169014},
        yticklabels={\hphantom{0}-500, \hphantom{-000}0, \hphantom{-0}500, \hphantom{-00}1k, \hphantom{-}1.5k, \hphantom{-00}2k, \hphantom{-}2.5k, \hphantom{-00}3k, \hphantom{-}3.5k, \hphantom{-00}4k, \hphantom{-}4.5k},
    ]
    \addplot graphics [
    ymin=-0.4650907377367773, ymax=10.005890429168739,
    xmin=-10.0, xmax=210.0,
    ]{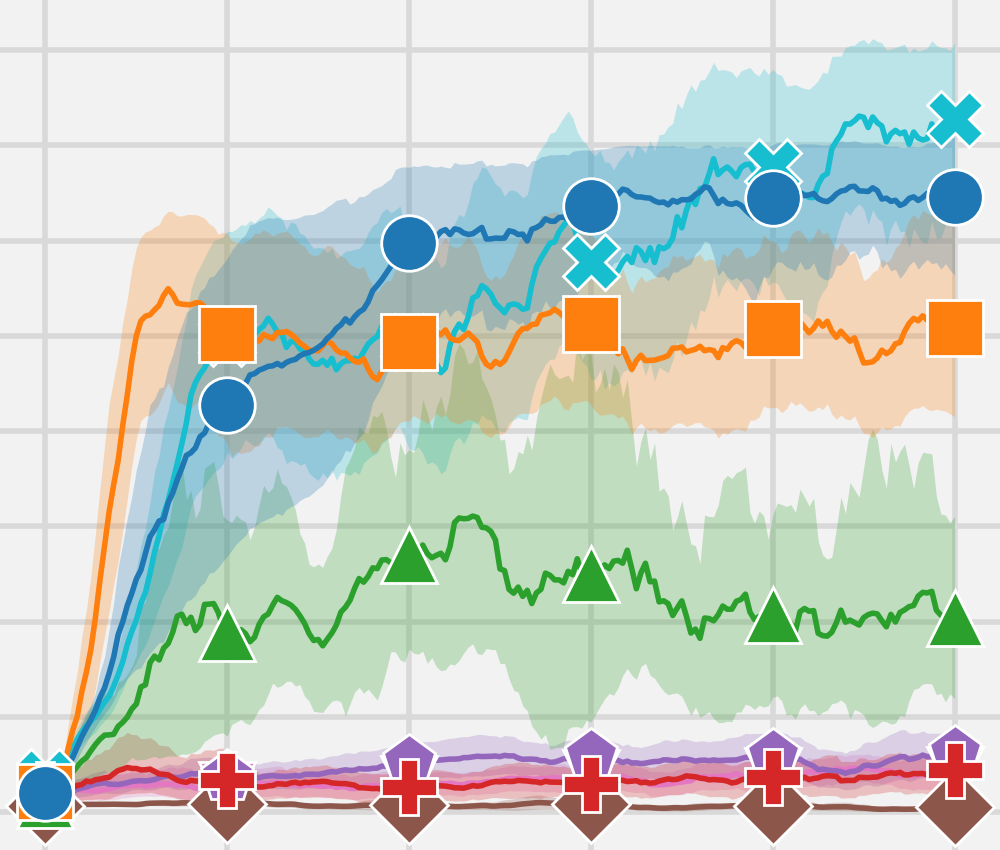};
    \end{axis}
\end{tikzpicture}
\begin{tikzpicture}[trim axis right]
    \begin{axis}[
        height=0.1615\textwidth,
        width=0.19\textwidth,
        title={\vphantom{p}Humanoid},
        ylabel={},
        xlabel={Time steps (1M)},
        xtick={0.0, 40.0, 80.0, 120.0, 160.0, 200.0},
        xticklabels={0.0, 0.2, 0.4, 0.6, 0.8, 1.0},
        ytick={-1.7809439002671417, 0.0, 1.7809439002671417, 3.5618878005342833, 5.342831700801424, 7.123775601068567, 8.904719501335707, 10.685663401602849},
        yticklabels={\hphantom{00}-2k, \hphantom{-000}0, \hphantom{-00}2k, \hphantom{-00}4k, \hphantom{-00}6k, \hphantom{-00}8k, \hphantom{-0}10k, \hphantom{-0}12k},
    ]
    \addplot graphics [
    ymin=-0.38895717946584696, ymax=10.00337858205326,
    xmin=-10.0, xmax=210.0,
    ]{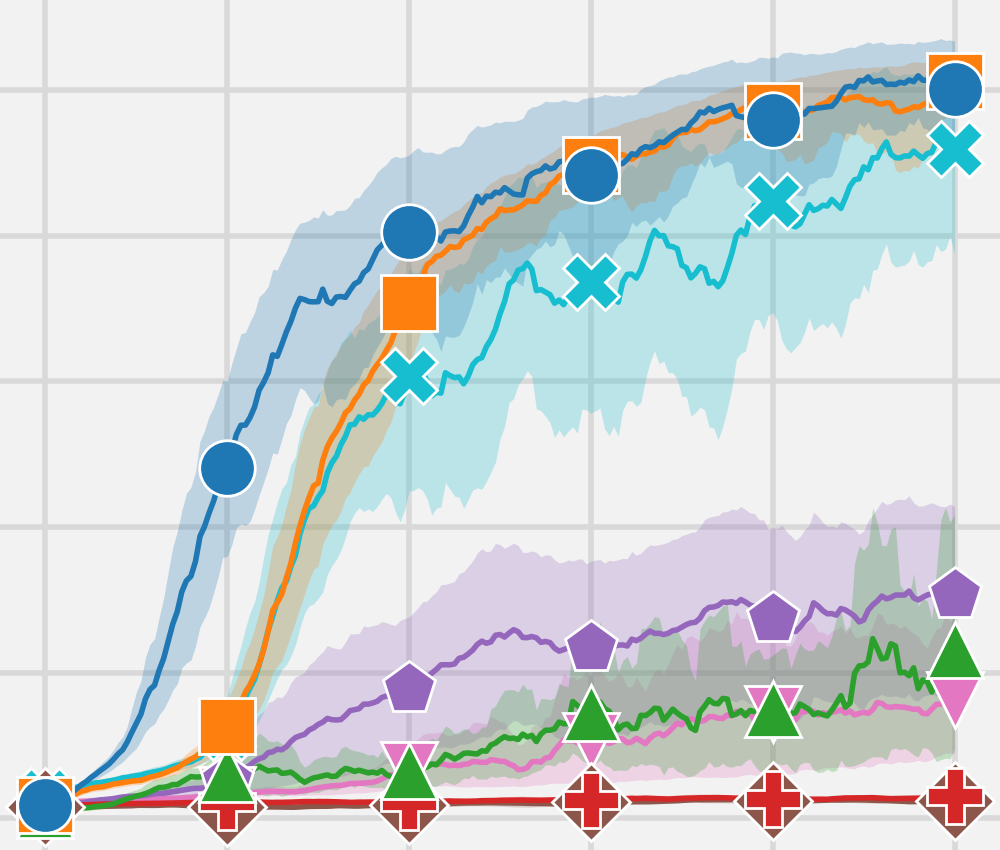};
    \end{axis}
\end{tikzpicture}

\begin{tikzpicture}[trim axis right]
    \begin{axis}[
        height=0.1615\textwidth,
        width=0.19\textwidth,
        title={\vphantom{p}Walker2d},
        ylabel={Total Reward},
        xlabel={Time steps (1M)},
        xtick={0.0, 40.0, 80.0, 120.0, 160.0, 200.0},
        xticklabels={0.0, 0.2, 0.4, 0.6, 0.8, 1.0},
        ytick={-1.5015015015015014, 0.0, 1.5015015015015014, 3.003003003003003, 4.504504504504505, 6.006006006006006, 7.5075075075075075, 9.00900900900901, 10.51051051051051},
        yticklabels={\hphantom{00}-1k, \hphantom{-000}0, \hphantom{-00}1k, \hphantom{-00}2k, \hphantom{-00}3k, \hphantom{-00}4k, \hphantom{-00}5k, \hphantom{-00}6k, \hphantom{-00}7k},
    ]
    \addplot graphics [
    ymin=-0.48343417948175965, ymax=10.01327874876635,
    xmin=-10.0, xmax=210.0,
    ]{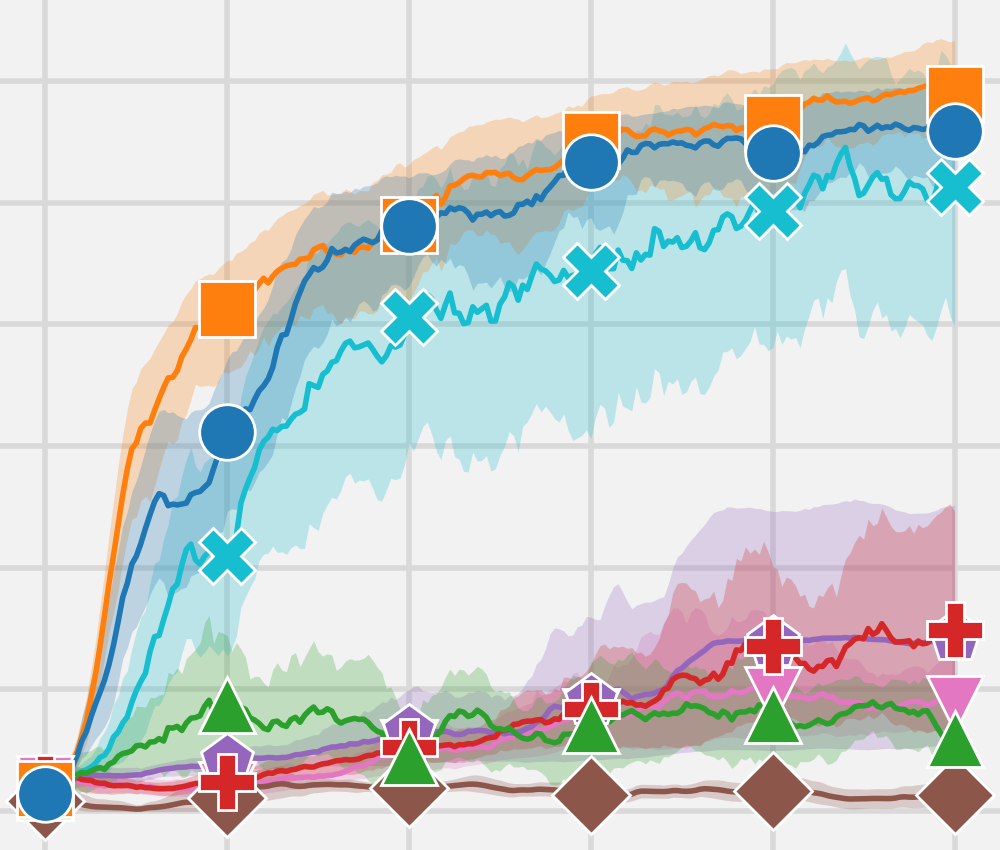};
    \end{axis}
\end{tikzpicture}
        
\centering

\fcolorbox{gray!10}{gray!10}{
\small
\colorcircle{white}{sb_blue} MRS.Q (Ours) \quad
\colorsquare{white}{sb_orange} MR.Q \quad
\colortriangle{white}{sb_green} MR.Q + MPC \quad
\colorplus{white}{sb_red} TD-MPC2 \quad
\colorpoly{white}{sb_purple} TD-M(PC)$^2$ \quad
\colordiamond{white}{sb_brown}  BMPC \quad
\colorinvertedtriangle{white}{sb_pink} BOOM \quad
\colorx{white}{sb_cyan} SimbaV2 
}

\caption{\textbf{Gym learning curves}. The shaded area captures a 95\% stratified bootstrap confidence interval across 10 seeds.}

\end{figure}

\clearpage

\subsection{DMC}

\begin{table}[ht]
\centering
\scriptsize
\caption{\textbf{DMC results.} Average final performance at 1M time steps. \textcolor{gray}{[Bracketed values]} indicate 95\% stratified bootstrap confidence intervals over 10 seeds.}
\setlength{\tabcolsep}{0pt}
\begin{tabularx}{\textwidth}{l@{\hspace{2pt}} rX@{\hspace{2pt}} rX@{\hspace{2pt}} rX@{\hspace{2pt}} rX@{\hspace{2pt}} rX@{\hspace{2pt}} rX@{\hspace{2pt}} rX@{\hspace{2pt}} rX}
\toprule
Task & \multicolumn{2}{c}{MRS.Q} & \multicolumn{2}{c}{MR.Q} & \multicolumn{2}{c}{MR.Q+MPC} & \multicolumn{2}{c}{TD-MPC2} & \multicolumn{2}{c}{TD-M(PC)$^2$} & \multicolumn{2}{c}{BMPC} & \multicolumn{2}{c}{BOOM} & \multicolumn{2}{c}{SimbaV2} \\
\midrule
acrobot-swingup & 468 & ~\textcolor{gray}{[417, 515]} & 567 & ~\textcolor{gray}{[517, 621]} & 459 & ~\textcolor{gray}{[264, 598]} & 554 & ~\textcolor{gray}{[391, 690]} & 592 & ~\textcolor{gray}{[550, 630]} & 550 & ~\textcolor{gray}{[503, 599]} & 619 & ~\textcolor{gray}{[600, 638]} & 497 & ~\textcolor{gray}{[424, 578]} \\
ball\_in\_cup-catch & 979 & ~\textcolor{gray}{[977, 982]} & 981 & ~\textcolor{gray}{[979, 983]} & 976 & ~\textcolor{gray}{[972, 982]} & 979 & ~\textcolor{gray}{[973, 986]} & 984 & ~\textcolor{gray}{[982, 986]} & 983 & ~\textcolor{gray}{[981, 986]} & 984 & ~\textcolor{gray}{[982, 986]} & 982 & ~\textcolor{gray}{[981, 985]} \\
cartpole-balance & 995 & ~\textcolor{gray}{[992, 998]} & 999 & ~\textcolor{gray}{[999, 1000]} & 974 & ~\textcolor{gray}{[924, 1000]} & 998 & ~\textcolor{gray}{[998, 999]} & 996 & ~\textcolor{gray}{[993, 998]} & 993 & ~\textcolor{gray}{[987, 998]} & 996 & ~\textcolor{gray}{[995, 999]} & 999 & ~\textcolor{gray}{[1000, 1000]} \\
cartpole-balance\_sparse & 1000 & ~\textcolor{gray}{[1000, 1000]} & 1000 & ~\textcolor{gray}{[1000, 1000]} & 1000 & ~\textcolor{gray}{[1000, 1000]} & 970 & ~\textcolor{gray}{[912, 1000]} & 999 & ~\textcolor{gray}{[998, 1000]} & 997 & ~\textcolor{gray}{[993, 1000]} & 1000 & ~\textcolor{gray}{[1000, 1000]} & 935 & ~\textcolor{gray}{[807, 1000]} \\
\bottomrule
\end{tabularx}
\end{table}

\input{figures/dmc_full_figure}

\clearpage

\subsection{HumanoidBench (No Hand)}

\begin{table}[ht]
\centering
\scriptsize
\caption{\textbf{HumanoidBench (No Hand) results.} Average final performance at 1M time steps. \textcolor{gray}{[Bracketed values]} indicate 95\% stratified bootstrap confidence intervals over 10 seeds. \texttt{h1-reach} results are not used in aggregate metrics.}
\setlength{\tabcolsep}{0pt}
\begin{tabularx}{\textwidth}{l@{\hspace{2pt}} rX@{\hspace{2pt}} rX@{\hspace{2pt}} rX@{\hspace{2pt}} rX@{\hspace{2pt}} rX@{\hspace{2pt}} rX@{\hspace{2pt}} rX@{\hspace{2pt}} rX}
\toprule
Task & \multicolumn{2}{c}{MRS.Q} & \multicolumn{2}{c}{MR.Q} & \multicolumn{2}{c}{MR.Q+MPC} & \multicolumn{2}{c}{TD-MPC2} & \multicolumn{2}{c}{TD-M(PC)$^2$} & \multicolumn{2}{c}{BMPC} & \multicolumn{2}{c}{BOOM} & \multicolumn{2}{c}{SimbaV2} \\
\midrule
h1-balance\_hard & 63 & ~\textcolor{gray}{[61, 66]} & 69 & ~\textcolor{gray}{[65, 76]} & 62 & ~\textcolor{gray}{[59, 65]} & 56 & ~\textcolor{gray}{[53, 60]} & 46 & ~\textcolor{gray}{[44, 49]} & 59 & ~\textcolor{gray}{[58, 61]} & 53 & ~\textcolor{gray}{[52, 56]} & 136 & ~\textcolor{gray}{[131, 143]} \\
h1-balance\_simple & 475 & ~\textcolor{gray}{[380, 557]} & 240 & ~\textcolor{gray}{[208, 275]} & 214 & ~\textcolor{gray}{[168, 266]} & 397 & ~\textcolor{gray}{[373, 424]} & 121 & ~\textcolor{gray}{[63, 206]} & 332 & ~\textcolor{gray}{[246, 425]} & 373 & ~\textcolor{gray}{[289, 435]} & 378 & ~\textcolor{gray}{[255, 472]} \\
h1-crawl & 966 & ~\textcolor{gray}{[961, 969]} & 931 & ~\textcolor{gray}{[913, 949]} & 777 & ~\textcolor{gray}{[767, 788]} & 968 & ~\textcolor{gray}{[964, 972]} & 799 & ~\textcolor{gray}{[737, 849]} & 795 & ~\textcolor{gray}{[664, 932]} & 795 & ~\textcolor{gray}{[664, 932]} & 872 & ~\textcolor{gray}{[814, 932]} \\
h1-hurdle & 182 & ~\textcolor{gray}{[165, 205]} & 128 & ~\textcolor{gray}{[111, 146]} & 214 & ~\textcolor{gray}{[184, 240]} & 387 & ~\textcolor{gray}{[293, 483]} & 170 & ~\textcolor{gray}{[155, 185]} & 204 & ~\textcolor{gray}{[187, 222]} & 412 & ~\textcolor{gray}{[337, 501]} & 61 & ~\textcolor{gray}{[50, 74]} \\
h1-maze & 324 & ~\textcolor{gray}{[319, 330]} & 345 & ~\textcolor{gray}{[338, 353]} & 303 & ~\textcolor{gray}{[287, 317]} & 232 & ~\textcolor{gray}{[170, 291]} & 331 & ~\textcolor{gray}{[331, 333]} & 291 & ~\textcolor{gray}{[243, 319]} & 226 & ~\textcolor{gray}{[168, 285]} & 335 & ~\textcolor{gray}{[329, 343]} \\
\bottomrule
\end{tabularx}
\end{table}

\input{figures/hb_full_figure}

\clearpage

\subsection{HumanoidBench (Hand)}

\begin{table}[ht]
\centering
\scriptsize
\caption{\textbf{HumanoidBench (Hand) results.} Average final performance at 1M time steps. \textcolor{gray}{[Bracketed values]} indicate 95\% stratified bootstrap confidence intervals over 10 seeds. \texttt{h1hand-reach} results are not used in aggregate metrics.}
\setlength{\tabcolsep}{0pt}
\begin{tabularx}{\textwidth}{l@{\hspace{1pt}} rX@{\hspace{1pt}} rX@{\hspace{1pt}} rX@{\hspace{1pt}} rX@{\hspace{1pt}} rX@{\hspace{1pt}} rX@{\hspace{1pt}} rX@{\hspace{1pt}} rX}
\toprule
Task & \multicolumn{2}{c}{MRS.Q} & \multicolumn{2}{c}{MR.Q} & \multicolumn{2}{c}{MR.Q+MPC} & \multicolumn{2}{c}{TD-MPC2} & \multicolumn{2}{c}{TD-M(PC)$^2$} & \multicolumn{2}{c}{BMPC} & \multicolumn{2}{c}{BOOM} & \multicolumn{2}{c}{SimbaV2} \\
\midrule
h1hand-balance\_hard & 80 & ~\textcolor{gray}{[77, 84]} & 79 & ~\textcolor{gray}{[76, 82]} & 73 & ~\textcolor{gray}{[67, 80]} & 66 & ~\textcolor{gray}{[59, 73]} & 64 & ~\textcolor{gray}{[60, 70]} & 69 & ~\textcolor{gray}{[67, 72]} & 69 & ~\textcolor{gray}{[66, 73]} & 69 & ~\textcolor{gray}{[61, 76]} \\
h1hand-balance\_simple & 407 & ~\textcolor{gray}{[311, 493]} & 143 & ~\textcolor{gray}{[123, 162]} & 234 & ~\textcolor{gray}{[190, 286]} & 93 & ~\textcolor{gray}{[67, 125]} & 100 & ~\textcolor{gray}{[85, 116]} & 278 & ~\textcolor{gray}{[143, 423]} & 87 & ~\textcolor{gray}{[78, 98]} & 93 & ~\textcolor{gray}{[87, 101]} \\
h1hand-crawl & 898 & ~\textcolor{gray}{[897, 901]} & 775 & ~\textcolor{gray}{[734, 809]} & 680 & ~\textcolor{gray}{[620, 724]} & 958 & ~\textcolor{gray}{[933, 978]} & 958 & ~\textcolor{gray}{[933, 978]} & 717 & ~\textcolor{gray}{[574, 822]} & 717 & ~\textcolor{gray}{[574, 822]} & 568 & ~\textcolor{gray}{[539, 602]} \\
h1hand-hurdle & 128 & ~\textcolor{gray}{[121, 135]} & 50 & ~\textcolor{gray}{[35, 68]} & 103 & ~\textcolor{gray}{[91, 115]} & 44 & ~\textcolor{gray}{[25, 66]} & 179 & ~\textcolor{gray}{[176, 184]} & 127 & ~\textcolor{gray}{[92, 165]} & 50 & ~\textcolor{gray}{[42, 58]} & 20 & ~\textcolor{gray}{[16, 25]} \\
h1hand-maze & 294 & ~\textcolor{gray}{[268, 312]} & 274 & ~\textcolor{gray}{[245, 300]} & 250 & ~\textcolor{gray}{[223, 277]} & 136 & ~\textcolor{gray}{[111, 180]} & 136 & ~\textcolor{gray}{[111, 180]} & 317 & ~\textcolor{gray}{[300, 338]} & 284 & ~\textcolor{gray}{[164, 355]} & 244 & ~\textcolor{gray}{[174, 315]} \\
\bottomrule
\end{tabularx}
\end{table}

\input{figures/hb_hand_full_figure}

\clearpage

\section{Ablation Results}

\begin{table}[ht]
\centering
\tiny
\caption{\textbf{Gym ablation results for MRS.Q.} Average final performance at 1M time steps. \textcolor{gray}{[Bracketed values]} indicate 95\% stratified bootstrap confidence intervals over 5 seeds.}
\setlength{\tabcolsep}{0pt}
\begin{tabularx}{\textwidth}{l@{\hspace{4pt}} rX@{\hspace{4pt}} rX@{\hspace{4pt}} rX@{\hspace{4pt}} rX@{\hspace{4pt}} rX@{\hspace{4pt}} rX@{\hspace{4pt}} rX@{\hspace{4pt}} rX}
\toprule
Task & \multicolumn{2}{c}{MRS.Q} & \multicolumn{2}{c}{2 VF} & \multicolumn{2}{c}{5 VF} & \multicolumn{2}{c}{20 VF} & \multicolumn{2}{c}{Exploration} & \multicolumn{2}{c}{SEM} & \multicolumn{2}{c}{Min (Ensemble)} & \multicolumn{2}{c}{Min (MPC)} \\
\midrule
Ant & 7189 & ~\textcolor{gray}{[6394, 7804]} & 6275 & ~\textcolor{gray}{[4468, 7622]} & 6121 & ~\textcolor{gray}{[4723, 7267]} & 6260 & ~\textcolor{gray}{[5578, 7140]} & 6822 & ~\textcolor{gray}{[6265, 7283]} & 4743 & ~\textcolor{gray}{[2739, 6749]} & 5789 & ~\textcolor{gray}{[4324, 7134]} & 5682 & ~\textcolor{gray}{[4282, 7091]} \\
HalfCheetah & 14649 & ~\textcolor{gray}{[14149, 15158]} & 14778 & ~\textcolor{gray}{[14454, 15158]} & 11574 & ~\textcolor{gray}{[7195, 14705]} & 13751 & ~\textcolor{gray}{[13477, 14209]} & 12438 & ~\textcolor{gray}{[12155, 12724]} & 12791 & ~\textcolor{gray}{[12254, 13329]} & 14355 & ~\textcolor{gray}{[14066, 14659]} & 13137 & ~\textcolor{gray}{[13138, 13138]} \\
Hopper & 3248 & ~\textcolor{gray}{[2945, 3516]} & 1538 & ~\textcolor{gray}{[1169, 1855]} & 2874 & ~\textcolor{gray}{[2225, 3415]} & 3233 & ~\textcolor{gray}{[2760, 3526]} & 2866 & ~\textcolor{gray}{[1751, 3468]} & 3131 & ~\textcolor{gray}{[2699, 3564]} & 1498 & ~\textcolor{gray}{[1101, 1855]} & 1899 & ~\textcolor{gray}{[961, 2969]} \\
Humanoid & 10526 & ~\textcolor{gray}{[10186, 10816]} & 4425 & ~\textcolor{gray}{[2571, 5506]} & 7366 & ~\textcolor{gray}{[5766, 8769]} & 10486 & ~\textcolor{gray}{[10206, 10742]} & 9933 & ~\textcolor{gray}{[9292, 10513]} & 8912 & ~\textcolor{gray}{[7617, 10208]} & 8418 & ~\textcolor{gray}{[6622, 10214]} & 2730 & ~\textcolor{gray}{[2158, 3461]} \\
\bottomrule
\end{tabularx}
\end{table}

\begin{table}[ht]
\centering
\tiny
\caption{\textbf{Gym ablation results for TD-MPC2.} Average final performance at 1M time steps. \textcolor{gray}{[Bracketed values]} indicate 95\% stratified bootstrap confidence intervals over 10 seeds.}
\setlength{\tabcolsep}{0pt}
\begin{tabular}{l@{\hspace{12pt}} rl@{\hspace{12pt}} rl}
\toprule
Task & & TD-MPC2 & & Min (10) \\
\midrule
Ant & 1938 & ~\textcolor{gray}{[795, 3694]} & 4412 & ~\textcolor{gray}{[3008, 5963]} \\
HalfCheetah & 11188 & ~\textcolor{gray}{[6871, 15475]} & 9024 & ~\textcolor{gray}{[2649, 15400]} \\
Hopper & 267 & ~\textcolor{gray}{[170, 360]} & 237 & ~\textcolor{gray}{[193, 265]} \\
Humanoid & 297 & ~\textcolor{gray}{[260, 336]} & 319 & ~\textcolor{gray}{[218, 418]} \\
Walker2d & 1537 & ~\textcolor{gray}{[668, 2592]} & 1236 & ~\textcolor{gray}{[888, 1629]} \\
\midrule
Mean & 0.41 & ~\textcolor{gray}{[0.27, 0.57]} & 0.48 & ~\textcolor{gray}{[0.33, 0.62]} \\
Median & 0.39 & ~\textcolor{gray}{[0.17, 0.65]} & 0.31 & ~\textcolor{gray}{[0.22, 0.40]} \\
IQM & 0.32 & ~\textcolor{gray}{[0.15, 0.53]} & 0.41 & ~\textcolor{gray}{[0.21, 0.54]} \\
\bottomrule
\end{tabular}
\end{table}

\begin{table}[ht]
\centering
\scriptsize
\caption{\textbf{DMC ablation results for MRS.Q.} Average final performance at 1M time steps. \textcolor{gray}{[Bracketed values]} indicate 95\% stratified bootstrap confidence intervals over 5 seeds.}
\setlength{\tabcolsep}{0pt}
\begin{tabularx}{\textwidth}{l@{\hspace{1pt}} rX@{\hspace{1pt}} rX@{\hspace{1pt}} rX@{\hspace{1pt}} rX@{\hspace{1pt}} rX@{\hspace{1pt}} rX@{\hspace{1pt}} rX@{\hspace{1pt}} rX}
\toprule
Task & \multicolumn{2}{c}{MRS.Q} & \multicolumn{2}{c}{2 VF} & \multicolumn{2}{c}{5 VF} & \multicolumn{2}{c}{20 VF} & \multicolumn{2}{c}{Exploration} & \multicolumn{2}{c}{SEM} & \multicolumn{2}{c}{Min (Ensemble)} & \multicolumn{2}{c}{Min (MPC)} \\
\midrule
acrobot-swingup & 468 & ~\textcolor{gray}{[417, 515]} & 604 & ~\textcolor{gray}{[561, 646]} & 516 & ~\textcolor{gray}{[438, 583]} & 304 & ~\textcolor{gray}{[231, 378]} & 429 & ~\textcolor{gray}{[307, 520]} & 496 & ~\textcolor{gray}{[311, 658]} & 647 & ~\textcolor{gray}{[605, 691]} & 644 & ~\textcolor{gray}{[614, 674]} \\
ball\_in\_cup-catch & 979 & ~\textcolor{gray}{[977, 982]} & 974 & ~\textcolor{gray}{[973, 976]} & 979 & ~\textcolor{gray}{[977, 982]} & 965 & ~\textcolor{gray}{[943, 978]} & 977 & ~\textcolor{gray}{[976, 978]} & 975 & ~\textcolor{gray}{[975, 977]} & 976 & ~\textcolor{gray}{[975, 977]} & 976 & ~\textcolor{gray}{[975, 978]} \\
cartpole-balance & 995 & ~\textcolor{gray}{[992, 998]} & 996 & ~\textcolor{gray}{[993, 998]} & 997 & ~\textcolor{gray}{[997, 999]} & 995 & ~\textcolor{gray}{[993, 998]} & 998 & ~\textcolor{gray}{[997, 999]} & 997 & ~\textcolor{gray}{[995, 999]} & 996 & ~\textcolor{gray}{[995, 998]} & 997 & ~\textcolor{gray}{[996, 998]} \\
cartpole-balance\_sparse & 1000 & ~\textcolor{gray}{[1000, 1000]} & 1000 & ~\textcolor{gray}{[1000, 1000]} & 1000 & ~\textcolor{gray}{[1000, 1000]} & 1000 & ~\textcolor{gray}{[1000, 1000]} & 1000 & ~\textcolor{gray}{[1000, 1000]} & 1000 & ~\textcolor{gray}{[1000, 1000]} & 1000 & ~\textcolor{gray}{[1000, 1000]} & 1000 & ~\textcolor{gray}{[1000, 1000]} \\
\bottomrule
\end{tabularx}
\end{table}

\begin{table}[ht]
\centering
\scriptsize
\caption{\textbf{DMC ablation results for TD-MPC2.} Average final performance at 1M time steps. \textcolor{gray}{[Bracketed values]} indicate 95\% stratified bootstrap confidence intervals over 10 seeds.}
\setlength{\tabcolsep}{0pt}
\begin{tabular}{l@{\hspace{12pt}} rl@{\hspace{12pt}} rl}
\toprule
Task & & TD-MPC2 & & Min (10) \\
\midrule
acrobot-swingup & 554 & ~\textcolor{gray}{[391, 690]} & 489 & ~\textcolor{gray}{[281, 633]} \\
ball\_in\_cup-catch & 979 & ~\textcolor{gray}{[973, 986]} & 984 & ~\textcolor{gray}{[982, 986]} \\
cartpole-balance & 998 & ~\textcolor{gray}{[998, 999]} & 998 & ~\textcolor{gray}{[996, 1000]} \\
cartpole-balance\_sparse & 970 & ~\textcolor{gray}{[912, 1000]} & 1000 & ~\textcolor{gray}{[1000, 1000]} \\
cartpole-swingup & 879 & ~\textcolor{gray}{[875, 882]} & 877 & ~\textcolor{gray}{[871, 882]} \\
cartpole-swingup\_sparse & 676 & ~\textcolor{gray}{[425, 849]} & 850 & ~\textcolor{gray}{[850, 850]} \\
cheetah-run & 915 & ~\textcolor{gray}{[913, 918]} & 886 & ~\textcolor{gray}{[864, 908]} \\
\bottomrule
\end{tabular}
\end{table}

\begin{table}[ht]
\centering
\scriptsize
\caption{\textbf{HumanoidBench (No Hand) ablation results for MRS.Q.} Average final performance at 1M time steps. \textcolor{gray}{[Bracketed values]} indicate 95\% stratified bootstrap confidence intervals over 5 seeds. \texttt{h1-reach} results are not used in aggregate metrics.}
\setlength{\tabcolsep}{0pt}
\begin{tabularx}{\textwidth}{l@{\hspace{1pt}} rX@{\hspace{1pt}} rX@{\hspace{1pt}} rX@{\hspace{1pt}} rX@{\hspace{1pt}} rX@{\hspace{1pt}} rX@{\hspace{1pt}} rX@{\hspace{1pt}} rX}
\toprule
Task & \multicolumn{2}{c}{MRS.Q} & \multicolumn{2}{c}{2 VF} & \multicolumn{2}{c}{5 VF} & \multicolumn{2}{c}{20 VF} & \multicolumn{2}{c}{Exploration} & \multicolumn{2}{c}{SEM} & \multicolumn{2}{c}{Min (Ensemble)} & \multicolumn{2}{c}{Min (MPC)} \\
\midrule
h1-balance\_hard & 63 & ~\textcolor{gray}{[61, 66]} & 58 & ~\textcolor{gray}{[52, 65]} & 64 & ~\textcolor{gray}{[61, 68]} & 65 & ~\textcolor{gray}{[56, 75]} & 56 & ~\textcolor{gray}{[56, 58]} & 67 & ~\textcolor{gray}{[63, 72]} & 61 & ~\textcolor{gray}{[55, 67]} & 61 & ~\textcolor{gray}{[57, 67]} \\
h1-balance\_simple & 475 & ~\textcolor{gray}{[380, 557]} & 278 & ~\textcolor{gray}{[218, 338]} & 532 & ~\textcolor{gray}{[470, 608]} & 569 & ~\textcolor{gray}{[493, 653]} & 303 & ~\textcolor{gray}{[264, 346]} & 512 & ~\textcolor{gray}{[361, 600]} & 430 & ~\textcolor{gray}{[344, 517]} & 491 & ~\textcolor{gray}{[393, 558]} \\
h1-crawl & 966 & ~\textcolor{gray}{[961, 969]} & 841 & ~\textcolor{gray}{[760, 914]} & 954 & ~\textcolor{gray}{[948, 960]} & 966 & ~\textcolor{gray}{[961, 971]} & 968 & ~\textcolor{gray}{[965, 972]} & 971 & ~\textcolor{gray}{[970, 973]} & 948 & ~\textcolor{gray}{[928, 969]} & 949 & ~\textcolor{gray}{[921, 964]} \\
h1-hurdle & 182 & ~\textcolor{gray}{[165, 205]} & 86 & ~\textcolor{gray}{[75, 100]} & 177 & ~\textcolor{gray}{[156, 199]} & 205 & ~\textcolor{gray}{[193, 220]} & 185 & ~\textcolor{gray}{[173, 200]} & 414 & ~\textcolor{gray}{[415, 415]} & 122 & ~\textcolor{gray}{[104, 140]} & 135 & ~\textcolor{gray}{[116, 155]} \\
h1-maze & 324 & ~\textcolor{gray}{[319, 330]} & 297 & ~\textcolor{gray}{[283, 316]} & 320 & ~\textcolor{gray}{[309, 331]} & 322 & ~\textcolor{gray}{[313, 332]} & 319 & ~\textcolor{gray}{[316, 325]} & 356 & ~\textcolor{gray}{[356, 356]} & 318 & ~\textcolor{gray}{[315, 324]} & 312 & ~\textcolor{gray}{[297, 324]} \\
\bottomrule
\end{tabularx}
\end{table}

\begin{table}[ht]
\centering
\scriptsize
\caption{\textbf{HumanoidBench (No Hand) results for TD-MPC2.} Average final performance at 1M time steps. \textcolor{gray}{[Bracketed values]} indicate 95\% stratified bootstrap confidence intervals over 10 seeds. \texttt{h1-reach} results are not used in aggregate metrics.}
\setlength{\tabcolsep}{0pt}
\begin{tabular}{l@{\hspace{12pt}} rl@{\hspace{12pt}} rl}
\toprule
Task & & TD-MPC2 & & Min (10) \\
\midrule
h1-balance\_hard & 56 & ~\textcolor{gray}{[53, 60]} & 52 & ~\textcolor{gray}{[51, 54]} \\
h1-balance\_simple & 397 & ~\textcolor{gray}{[373, 424]} & 353 & ~\textcolor{gray}{[321, 394]} \\
h1-crawl & 968 & ~\textcolor{gray}{[964, 972]} & 969 & ~\textcolor{gray}{[968, 971]} \\
h1-hurdle & 387 & ~\textcolor{gray}{[293, 483]} & 365 & ~\textcolor{gray}{[308, 428]} \\
h1-maze & 232 & ~\textcolor{gray}{[170, 291]} & 216 & ~\textcolor{gray}{[157, 261]} \\
h1-pole & 730 & ~\textcolor{gray}{[583, 845]} & 851 & ~\textcolor{gray}{[762, 901]} \\
h1-reach & 7753 & ~\textcolor{gray}{[5514, 9314]} & 6943 & ~\textcolor{gray}{[6319, 7686]} \\
h1-run & 791 & ~\textcolor{gray}{[789, 793]} & 792 & ~\textcolor{gray}{[790, 794]} \\
h1-sit\_hard & 799 & ~\textcolor{gray}{[796, 803]} & 801 & ~\textcolor{gray}{[801, 802]} \\
h1-sit\_simple & 818 & ~\textcolor{gray}{[818, 819]} & 818 & ~\textcolor{gray}{[818, 820]} \\
h1-slide & 362 & ~\textcolor{gray}{[304, 414]} & 400 & ~\textcolor{gray}{[355, 452]} \\
h1-stair & 374 & ~\textcolor{gray}{[216, 538]} & 516 & ~\textcolor{gray}{[388, 615]} \\
h1-stand & 817 & ~\textcolor{gray}{[815, 819]} & 818 & ~\textcolor{gray}{[818, 820]} \\
h1-walk & 816 & ~\textcolor{gray}{[815, 817]} & 816 & ~\textcolor{gray}{[815, 818]} \\
\midrule
Mean & 580 & ~\textcolor{gray}{[560, 600]} & 598 & ~\textcolor{gray}{[583, 612]} \\
Median & 730 & ~\textcolor{gray}{[583, 793]} & 792 & ~\textcolor{gray}{[762, 794]} \\
IQM & 613 & ~\textcolor{gray}{[591, 648]} & 644 & ~\textcolor{gray}{[625, 666]} \\
\bottomrule
\end{tabular}
\end{table}

\clearpage

~
\vspace{36pt}

\begin{table}[ht]
\centering
\scriptsize
\caption{\textbf{HumanoidBench (Hand) ablation results for MRS.Q.} Average final performance at 1M time steps. \textcolor{gray}{[Bracketed values]} indicate 95\% stratified bootstrap confidence intervals over 5 seeds. \texttt{h1hand-reach} results are not used in aggregate metrics.}
\setlength{\tabcolsep}{0pt}
\begin{tabularx}{\textwidth}{l@{\hspace{1pt}} rX@{\hspace{1pt}} rX@{\hspace{1pt}} rX@{\hspace{1pt}} rX@{\hspace{1pt}} rX@{\hspace{1pt}} rX@{\hspace{1pt}} rX@{\hspace{1pt}} rX}
\toprule
Task & \multicolumn{2}{c}{MRS.Q} & \multicolumn{2}{c}{2 VF} & \multicolumn{2}{c}{5 VF} & \multicolumn{2}{c}{20 VF} & \multicolumn{2}{c}{Exploration} & \multicolumn{2}{c}{SEM} & \multicolumn{2}{c}{Min (Ensemble)} & \multicolumn{2}{c}{Min (MPC)} \\
\midrule
h1hand-balance\_hard & 80 & ~\textcolor{gray}{[77, 84]} & 70 & ~\textcolor{gray}{[65, 77]} & 83 & ~\textcolor{gray}{[77, 90]} & 79 & ~\textcolor{gray}{[75, 85]} & 82 & ~\textcolor{gray}{[80, 85]} & 76 & ~\textcolor{gray}{[75, 77]} & 83 & ~\textcolor{gray}{[78, 88]} & 81 & ~\textcolor{gray}{[77, 88]} \\
h1hand-balance\_simple & 407 & ~\textcolor{gray}{[311, 493]} & 107 & ~\textcolor{gray}{[90, 134]} & 344 & ~\textcolor{gray}{[256, 440]} & 465 & ~\textcolor{gray}{[343, 608]} & 224 & ~\textcolor{gray}{[160, 306]} & 635 & ~\textcolor{gray}{[635, 635]} & 309 & ~\textcolor{gray}{[249, 374]} & 330 & ~\textcolor{gray}{[242, 423]} \\
h1hand-crawl & 898 & ~\textcolor{gray}{[897, 901]} & 595 & ~\textcolor{gray}{[510, 691]} & 788 & ~\textcolor{gray}{[773, 806]} & 860 & ~\textcolor{gray}{[817, 911]} & 813 & ~\textcolor{gray}{[774, 853]} & 969 & ~\textcolor{gray}{[969, 969]} & 784 & ~\textcolor{gray}{[737, 836]} & 756 & ~\textcolor{gray}{[692, 835]} \\
h1hand-hurdle & 128 & ~\textcolor{gray}{[121, 135]} & 39 & ~\textcolor{gray}{[31, 49]} & 81 & ~\textcolor{gray}{[74, 88]} & 158 & ~\textcolor{gray}{[140, 177]} & 116 & ~\textcolor{gray}{[107, 127]} & 244 & ~\textcolor{gray}{[225, 270]} & 75 & ~\textcolor{gray}{[64, 84]} & 75 & ~\textcolor{gray}{[65, 84]} \\
\bottomrule
\end{tabularx}
\end{table}

\vspace{64pt}

\begin{table}[ht]
\centering
\scriptsize
\caption{\textbf{HumanoidBench (Hand) results for TD-MPC2.} Average final performance at 1M time steps. \textcolor{gray}{[Bracketed values]} indicate 95\% stratified bootstrap confidence intervals over 10 seeds. \texttt{h1hand-reach} results are not used in aggregate metrics.}
\setlength{\tabcolsep}{0pt}
\begin{tabular}{l@{\hspace{12pt}} rl@{\hspace{12pt}} rl}
\toprule
Task & & TD-MPC2 & & Min (10) \\
\midrule
h1hand-balance\_hard & 66 & ~\textcolor{gray}{[59, 73]} & 65 & ~\textcolor{gray}{[64, 68]} \\
h1hand-balance\_simple & 93 & ~\textcolor{gray}{[67, 125]} & 83 & ~\textcolor{gray}{[76, 93]} \\
h1hand-crawl & 958 & ~\textcolor{gray}{[933, 978]} & 903 & ~\textcolor{gray}{[862, 946]} \\
h1hand-hurdle & 44 & ~\textcolor{gray}{[25, 66]} & 42 & ~\textcolor{gray}{[12, 75]} \\
h1hand-maze & 136 & ~\textcolor{gray}{[111, 180]} & 142 & ~\textcolor{gray}{[123, 160]} \\
h1hand-pole & 139 & ~\textcolor{gray}{[103, 176]} & 235 & ~\textcolor{gray}{[191, 320]} \\
h1hand-reach & 6308 & ~\textcolor{gray}{[4024, 8276]} & 5338 & ~\textcolor{gray}{[4781, 5896]} \\
h1hand-run & 65 & ~\textcolor{gray}{[39, 97]} & 84 & ~\textcolor{gray}{[50, 133]} \\
h1hand-sit\_hard & 369 & ~\textcolor{gray}{[118, 628]} & 637 & ~\textcolor{gray}{[332, 852]} \\
h1hand-sit\_simple & 177 & ~\textcolor{gray}{[40, 395]} & 647 & ~\textcolor{gray}{[267, 921]} \\
h1hand-slide & 104 & ~\textcolor{gray}{[65, 146]} & 196 & ~\textcolor{gray}{[171, 220]} \\
h1hand-stair & 59 & ~\textcolor{gray}{[52, 68]} & 69 & ~\textcolor{gray}{[62, 79]} \\
h1hand-stand & 296 & ~\textcolor{gray}{[195, 394]} & 623 & ~\textcolor{gray}{[495, 752]} \\
h1hand-walk & 308 & ~\textcolor{gray}{[250, 365]} & 554 & ~\textcolor{gray}{[326, 783]} \\
\midrule
Mean & 216 & ~\textcolor{gray}{[193, 245]} & 329 & ~\textcolor{gray}{[287, 363]} \\
Median & 136 & ~\textcolor{gray}{[86, 152]} & 196 & ~\textcolor{gray}{[171, 216]} \\
IQM & 144 & ~\textcolor{gray}{[114, 168]} & 274 & ~\textcolor{gray}{[206, 286]} \\
\bottomrule\end{tabular}\end{table}

\end{document}